# Inference of Multiscale Gaussian Graphical Models

Edmond Sanou[1] Université Paris-Saclay, CNRS, Univ Evry, Laboratoire de Mathématiques et Modélisation d'Evry
Christophe Ambroise Université Paris-Saclay, CNRS, Univ Evry, Laboratoire de Mathématiques et Modélisation d'Evry
Geneviève Robin Université Paris-Saclay, CNRS, Univ Evry, Laboratoire de Mathématiques et Modélisation d'Evry

**Abstract**

Gaussian Graphical Models (GGMs) are widely used in high-dimensional data analysis to synthesize the interaction between variables. In many applications, such as genomics or image analysis, graphical models rely on sparsity and clustering to reduce dimensionality and improve performances. This paper explores a slightly different paradigm where clustering is not knowledge-driven but performed simultaneously with the graph inference task. We introduce a novel Multiscale Graphical Lasso (MGLasso) to improve networks interpretability by proposing graphs at different granularity levels. The method estimates clusters through a convex clustering approach — a relaxation of $k$-means, and hierarchical clustering. The conditional independence graph is simultaneously inferred through a neighborhood selection scheme for undirected graphical models. MGLasso extends and generalizes the sparse group fused lasso problem to undirected graphical models. We use continuation with Nesterov smoothing in a shrinkage-thresholding algorithm (CONESTA) to propose a regularization path of solutions along the group fused Lasso penalty, while the Lasso penalty is kept constant. Extensive experiments on synthetic data compare the performances of our model to state-of-the-art clustering methods and network inference models. Applications to gut microbiome data and poplar's methylation mixed with transcriptomic data are presented.

## Contents

[1]Corresponding author: doedmond.sanou@univ-evry.fr

# 1 Introduction

Probabilistic graphical models (Lauritzen 1996; Koller and Friedman 2009) are widely used in high-dimensional data analysis to synthesize the interaction between variables. In many applications, such as genomics or image analysis, graphical models reduce the number of parameters by selecting the most relevant interactions between variables. Undirected *Gaussian Graphical Models* (GGMs) are a class of graphical models used in Gaussian settings. In the context of high-dimensional statistics, graphical models are generally assumed sparse, meaning that a small number of variables interact compared to the total number of possible interactions. This assumption has been shown to provide both statistical and computational advantages by simplifying the structure of dependence between variables (Dempster 1972) and allowing efficient algorithms (Meinshausen and Bühlmann 2006). See, for instance, Fan, Liao, and Liu (2016) for a review of sparse graphical models inference.

In GGMs, it is well known (Lauritzen 1996) that inferring the graphical model or, equivalently, the *conditional independence graph* (CIG) boils down to inferring the support of the precision matrix (the inverse of the variance-covariance matrix). Several $\ell_1$ penalized methods have been proposed in the literature to learn the CIG of GGMs. For instance, *the neighborhood selection* (MB, Meinshausen and Bühlmann 2006) based on a nodewise regression approach via the *least absolute shrinkage and selection operator* (Lasso, R. Tibshirani 1996) is a popular method. Each variable is regressed on the others, taking advantage of the link between the so-obtained regression coefficients and partial correlations. The MB method has generated a long line of work in nodewise regression methods. For instance, Rocha, Zhao, and Yu (2008) and Ambroise, Chiquet, and Matias (2009) showed that nodewise regression could be seen as a pseudo-likelihood approximation and Peng et al. (2009) extended the MB method to estimate sparse partial correlations using a single regression problem. Other inference methods similar to nodewise regression include a method based on the Dantzig selector (Yuan 2010) and the introduction of the Clime estimator (Cai, Liu, and Luo 2011). Another family of sparse CIG inference methods directly estimates  via direct minimization of the $\ell_1$-penalized negative log-likelihood (Banerjee, El Ghaoui, and d'Aspremont 2008), without resorting to the auxiliary regression problem. This method called the *graphical Lasso* (GLasso, Friedman, Hastie, and Tibshirani 2007), benefits from many optimization algorithms (Yuan and Lin 2007; Rothman et al. 2008; Banerjee, El Ghaoui, and d'Aspremont 2008; Hsieh et al. 2014).

Such inference methods are widely used and enjoy many favorable theoretical and empirical properties, including robustness to high-dimensional problems. However, some limitations have been observed, particularly in the presence of strongly correlated variables. Known impairments of Lasso-type regularization cause these limitations in this context (Bühlmann et al. 2012; Park, Hastie, and Tibshirani 2006). To overcome this, in addition to sparsity, several previous works attempt to estimate CIG by integrating clustering structures among variables for statistical sanity and interpretability. A non-exhaustive list of works that integrate a clustering structure to speed up or improve the estimation procedure includes Honorio et al. (2009), Ambroise, Chiquet, and Matias (2009), Mazumder and Hastie (2012), Tan, Witten, and Shojaie (2013), Devijver and Gallopin (2018), Yao and Allen (2019).

The above methods exploit the group structure to simplify the graph inference problem and infer the CIG between single variables. Another question that has received less attention is the inference of the CIG between the groups of variables, i.e., between the meta-variables representative of the group structure. A recent work introducing inference of graphical models on multiple grouping levels is Cheng, Shan, and Kim (2017). They proposed inferring the CIG of gene data on two levels corresponding to genes and pathways, respectively. Note that pathways are considered as groups of functionally related genes known in advance. The inference is achieved by optimizing a penalized maximum likelihood that estimates a sparse network at both gene and group levels. Our work is also part of this dynamic. We introduce a penalty term allowing parsimonious networks to be built at different clustering levels. The main difference with the procedure of Cheng, Shan, and Kim (2017) is that we do not require prior knowledge of the group structure, which makes the problem significantly more complex. In addition, our method has the advantage of proposing CIGs at more than two levels of granularity.

We introduce the Multiscale Graphical Lasso (MGLasso), a novel method to estimate simultaneously a hierarchical clustering structure and graphical models depicting the conditional independence structure between clusters of variables at each level of the hierarchy. Our approach is based on neighborhood selection (Meinshausen and Bühlmann 2006) and considers an additional fused-Lasso type penalty for clustering (Pelckmans et al. 2005; Hocking et al. 2011; Lindsten, Ohlsson, and Ljung 2011).

The use of fusion penalties in Gaussian graphical model inference is a well-studied area. Some prior works on learning sparse GGMs with a fusion penalty term have focused on penalized likelihood. Among those, a line of works (Danaher, Wang, and Witten 2014; S. Yang et al. 2015) infers multiple graphs across several classes while assuming the observations belong to different known clusters. Another line of research (Honorio et al. 2009; Yao and Allen 2019; Lin et al. 2020) investigates fusion penalties for enforcing local constancy in the nodes of the inferred network. Variables belonging to the same clusters are thus more likely to share the same neighborhood. These ordinary likelihood-based models are computationally challenging compared to pseudo-likelihood approximations. The unpublished manuscript of Ganguly and Polonik (2014) introduces a fusion-like penalty in the neighborhood selection framework. However, the problem is solved in a node-wise regression fashion where the $p$ regressions problems are not combined.

Fusion penalties have also been used in simple regression problems (Robert Tibshirani et al. 2005) and multivariate regression analysis (multitask learning) with multiple outcomes (see, e.g., Chen et al. 2010; Degras 2021; Dondelinger, Mukherjee, and Initiative 2020; Hallac, Leskovec, and Boyd 2015; Chu et al. 2021). The defined penalties encourage fusion between predictors in simple regression, or outcomes that share similar model coefficients in multitask learning. Fusions can be formulated in a general form assuming no order on the variables as in convex clustering (Hoefling 2010; Petry, Flexeder, and Tutz 2011) or assuming the availability of prior information about clusters (Rudin, Osher, and Fatemi 1992; Hallac, Leskovec, and Boyd 2015).

The multitask learning framework can be extended to the learning of GGMs. Chiquet, Grandvalet, and Ambroise (2011) introduced a multitask inference for multiple graphical models when observations belong to different clusters. In MGLasso, the multitask learning framework is combined with a novel general fusion penalty to uncover clustering over variables. In the defined fusion term, we consider reordering the regression coefficients to match common predictors and symmetric coefficients. That results in enforcing the grouping property by encouraging variables belonging to the same cluster to have the same neighborhood. MGLasso exploits the multitask learning framework for GGMs inference coupled with a convex clustering problem over the nodes to infer multiscale networks and clusters simultaneously. To our knowledge, this is the first attempt in the literature of undirected GGMs. MGLasso can also be seen as an extension of sparse group fused Lasso for graphical models and be straightforwardly extended to probability distributions belonging to the exponential family (E. Yang et al. 2012). The MGLasso algorithm is implemented in the R package *mglasso* available at https://CRAN.R-project.org/package=mglasso. The remainder of this paper is organized as follows. In Section 2 and Section 3, we formally introduce the Multiscale Graphical Lasso and its optimization algorithm. Section 4 presents simulated and real data numerical results.

## 2 Multiscale Graphical Lasso

Let $\mathbf{X} = (X^1, \dots, X^p)^T$ be a $p$-dimensional Gaussian random vector, with mean vector $\boldsymbol{\mu} \in \mathbb{R}^p$ and positive definite covariance matrix $\in \mathbb{R}^{p \times p}$. Let $G = (V, E)$ be a graph encoding the conditional independence structure of the normal distribution $\mathcal{N}(\boldsymbol{\mu}, )$, where $V = \{1, \dots p\}$ is the set of vertices and $E$ the set of edges. The graph $G$ is uniquely determined by the support of the precision matrix $= {}^{-1}$ (Dempster 1972). Specifically, for any two vertices $i \neq j \in V$, the edge $(i, j)$ belongs to the set $E$ if and only if $\Omega_{ij} \neq 0$. On the contrary, if $\Omega_{ij} = 0$, the variables $X^i$ and $X^j$ are said to be independent conditionally to the remaining variables $X^{\setminus (i,j)}$. We note,

$$X^i \perp\!\!\!\perp X^j | X^{\setminus (i,j)} \Leftrightarrow \Omega_{ij} = 0.$$

Let $\boldsymbol{X} = \left(\boldsymbol{X}_1^T, \dots, \boldsymbol{X}_n^T\right)^T$ be the $n \times p$-dimensional data matrix composed of $n$ i.i.d samples of the Gaussian random vector $\mathbf{X}$. To perform graphical model inference, Meinshausen and Bühlmann (2006) consider $p$ separate linear regressions of the form:

$$\hat{\boldsymbol{\beta}}^i(\lambda) = \underset{\boldsymbol{\beta}^i \in \mathbb{R}^{p-1}}{\operatorname{argmin}} \frac{1}{n} \left\| \mathbf{X}^i - \mathbf{X}^{\setminus i} \boldsymbol{\beta}^i \right\|_2^2 + \lambda \left\| \boldsymbol{\beta}^i \right\|_1, \tag{1}$$

where $\lambda$ is a non-negative regularization parameter, $\mathbf{X}^{\setminus i}$ denotes the matrix $\mathbf{X}$ deprived of column $i$, $\boldsymbol{\beta}^i = (\beta^i_j)_{j \in \{1, \dots, p\} \setminus i}$ is a vector of $p - 1$ regression coefficients and $\|.\|_1$ is the $\ell_1$−norm. These Lasso regularized problems estimate the neighborhoods, one variable at a time. The final edge set estimates $\hat{E}$ can be deduced from the union of the estimated neighborhoods using an AND or OR rule (Meinshausen and Bühlmann (2006)). The MB approach is based on the central relationship between simple linear regression and precision matrix coefficients. It can be shown that $\beta^i_j = -\frac{\Omega_{ij}}{\Omega_{ii}}$ (Lauritzen 1996).

On the other hand, let us now consider the clustering analysis of the $p$ variables in $\mathbb{R}^n$. The convex clustering problem (Hocking et al. 2011; Lindsten, Ohlsson, and Ljung 2011; Pelckmans et al. 2005) is the minimization of the quantity

$$\frac{1}{2} \sum_{i=1}^{p} \left\| \boldsymbol{X}^i - \boldsymbol{\alpha}^i \right\|_2^2 + \lambda \sum_{i<j} w_{ij} \left\| \boldsymbol{\alpha}^i - \boldsymbol{\alpha}^j \right\|_q \tag{2}$$

with respect to the matrix $\boldsymbol{\alpha} \in \mathbb{R}^{p\times n}$, where $\lambda$ is a sparsity penalization parameter, $\{w_{ij}\}$ are symmetric positive weights, $\boldsymbol{\alpha}^i \in \mathbb{R}^n$ is the centroid to which $\boldsymbol{X}^i$ is assigned to, and $\|.\|_q$ is the $\ell_q$-norm on $\mathbb{R}^p$ with $q \geq 1$. Points $\boldsymbol{X}^i$ and $\boldsymbol{X}^j$ are assigned to the same cluster if $\hat{\boldsymbol{\alpha}}^i \approx \hat{\boldsymbol{\alpha}}^j$. The regularization path of solutions to problem in Equation 2 can be represented as a dendrogram. The path properties have been studied in Chi and Lange (2015) and Chiquet, Gutierrez, and Rigaill (2017), among others. Note that these approaches rely on geometric properties of matrix $\boldsymbol{X}$, and do not require any assumption on the distribution of the covariates.

We propose to combine the MB and convex clustering approaches. Specifically, the $p$ independent Lasso regressions of the MB approach are merged into a single optimization criterion where a convex clustering fusion penalty in $\ell_2$ is applied on the regression vectors considered as cluster centers. Namely, the *Multiscale Graphical Lasso* (MGLasso) pseudo-likelihood problem minimizes in a Gaussian framework the following quantity:

$$J_{\lambda_1,\lambda_2}(\boldsymbol{\beta};\mathbf{X}) = \frac{1}{2}\sum_{i=1}^{p}\left\|\mathbf{X}^i - \mathbf{X}^{\setminus i}\boldsymbol{\beta}^i\right\|_2^2 + \lambda_1\sum_{i=1}^{p}\left\|\boldsymbol{\beta}^i\right\|_1 + \lambda_2\sum_{i<j}\left\|\boldsymbol{\beta}^i - \tau_{ij}\boldsymbol{\beta}^j\right\|_2, \tag{3}$$

with respect to $\boldsymbol{\beta} := [\boldsymbol{\beta}^1, \ldots, \boldsymbol{\beta}^p] \in \mathbb{R}^{(p-1)\times p}$, where $\mathbf{X}^i \in \mathbb{R}^n$ denotes the $i$-th column of $\mathbf{X}$, $\lambda_1$ and $\lambda_2$ are penalization parameters, $\tau_{ij} \in \mathbb{R}^{(p-1)\times(p-1)}$ is a permutation matrix, which permutes the coefficients in the regression vector $\boldsymbol{\beta}^j$ such as

$$\left\|\boldsymbol{\beta}^i - \tau_{ij}\boldsymbol{\beta}^j\right\|_2 = \sqrt{\sum_{k\in\{1,\ldots,p\}\setminus\{i,j\}}(\beta_k^i - \beta_k^j)^2 + (\beta_j^i - \beta_i^j)^2},$$

as illustrated in Figure 1. The coefficient $\beta_k^i$ is to be read as the multiple regression coefficients of $\boldsymbol{X}^i$ on $\boldsymbol{X}^k$.

The MGLasso criterion can be seen as a multitask regression problem where the set of responses is identical to the set of predictors. The Lasso penalty term encourages sparsity in the estimated coefficients while the group-fused term encourages fusion in the regression vectors $\boldsymbol{\beta}^i$ and $\boldsymbol{\beta}^j$.

Let us illustrate by an example the effect of the fusion term in the proposed approach. Two variables $i$ and $j$ are in the same group when $\|\boldsymbol{\beta}^i - \tau_{ij}\boldsymbol{\beta}^j\|_2 \approx 0$. Considering a cluster $\mathscr{C}$ of $q$ variables, it is straightforward to show that $\forall(i,j) \in \mathscr{C}^2$, we have $\hat{\beta}_j^i = \beta_{\mathscr{C}}$, where $\beta_{\mathscr{C}}$ is a scalar. Thus the algorithm is likely to produce precision matrices with blocks of constant entries for a given value of $\lambda_2$, each block corresponding to a cluster. In the same vein as Park, Hastie, and Tibshirani (2006), a cluster composed of variables that share the same coefficients can be summarized by a representative variable.

A component-wise difference between two regression vectors without reordering the coefficients would not necesarily cluster variables which share the same neighborhood. The permutation $\tau_{ij}$ reoders coefficients in such a way that differences are taken between symmetric coeffecients and those corresponding to the same set of predictors. The model is thus likely to cluster together variables that share the same neighboring structure and encourages symmetric graph structures.

In practice, when external information about the clustering structure is available, the problem can be generalized into:

$$\min_{\boldsymbol{\beta}}\sum_{i=1}^{p}\frac{1}{2}\left\|\mathbf{X}^i - \mathbf{X}^{\setminus i}\boldsymbol{\beta}^i\right\|_2^2 + \lambda_1\sum_{i=1}^{p}\left\|\boldsymbol{\beta}^i\right\|_1 + \lambda_2\sum_{i<j}w_{ij}\left\|\boldsymbol{\beta}^i - \tau_{ij}\boldsymbol{\beta}^j\right\|_2, \tag{4}$$

where $w_{ij}$ is a positive weight. In the remainder of the paper, we will assume that $w_{ij} = 1$ for simplicity.

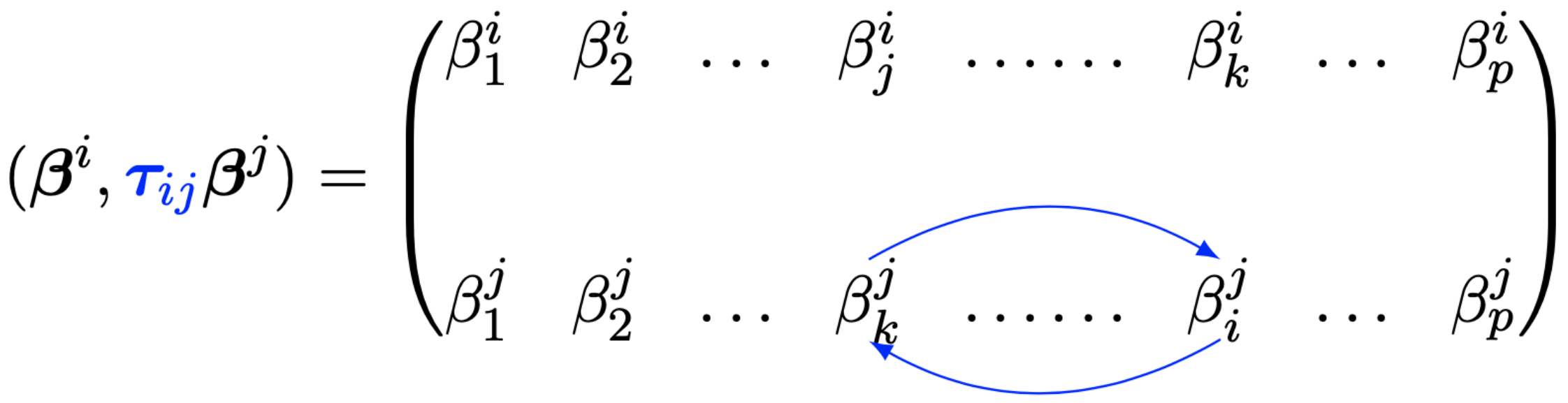

Figure 1: Illustration of the permutation between regression coefficients in the MGLasso model.

## 3 Numerical scheme

This Section introduces a complete numerical scheme of the Multiscale Graphical Lasso via convex optimization and a model selection procedure. Section 3.1 reviews the principles of the Continuation with Nesterov smoothing in a shrinkage-thresholding algorithm (CONESTA, Hadj-Selem et al. 2018). Section 3.2 details a reformulation of the MGLasso criterion, which eases the use of CONESTA as a solver. Finally, Section 3.3 presents the procedure for selecting the regularization parameters.

### 3.1 Optimization via CONESTA algorithm

The optimization problem for Multiscale Graphical Lasso is convex but not straightforward to solve using classical algorithms because of the fused-lasso type penalty, which is non-separable and admits no closed-form solution for the proximal gradient. We rely on the Continuation with Nesterov smoothing in a shrinkage-thresholding algorithm (Hadj-Selem et al. 2018) dedicated to high-dimensional regression problems with structured sparsity, such as group structures.

The CONESTA solver, initially introduced for neuro-imaging problems, addresses a general class of convex optimization problems that include group-wise penalties. The algorithm solves problems in the form

$$\text{minimize w. r. t.}\, \boldsymbol{\theta} \quad f(\boldsymbol{\theta}) = g(\boldsymbol{\theta}) + \lambda_1 h(\boldsymbol{\theta}) + \lambda_2 s(\boldsymbol{\theta}), \tag{5}$$

where $\boldsymbol{\theta} \in \mathbb{R}^d$ and $\lambda_1$ and $\lambda_2$ are penalty parameters.

In the original paper (Hadj-Selem et al. 2018), $g(\boldsymbol{\theta})$ is a differentiable function, $h(\boldsymbol{\theta})$ is a penalty function whose proximal operator $\text{prox}_{\lambda_1 h}$ is known in closed-form.

Given $\phi \subseteq \{1, \dots, d\}$, let $\boldsymbol{\theta}_\phi = (\theta_i)_{i \in \phi}$ denote the subvector of $\boldsymbol{\theta}$ referenced by the indices in $\phi$. Denote $\Phi = \{\phi_1, \dots, \phi_{\text{Card}(\Phi)}\}$ a collection with $\phi_i \subseteq \{1, \dots, d\}$. Let the matrix $\mathbf{A}_\phi \in \mathbb{R}^{m \times \text{Card}(\Phi)}$ define a linear map from $\mathbb{R}^{\text{Card}(\phi)}$ to $\mathbb{R}^m$ by sending the column vector $\boldsymbol{\theta}_\phi \in \mathbb{R}^{\text{Card}(\phi)}$ to the column vector $\mathbf{A}_\phi \boldsymbol{\theta}_\phi \in \mathbb{R}^m$. The function $s(\boldsymbol{\theta})$ is assumed to be an $\ell_{1,2}$-norm i.e., the sum of the group-wise $\ell_2$-norms of the elements $\mathbf{A}_\phi \boldsymbol{\theta}_\phi, \phi \in \Phi$. Namely,

$$s(\boldsymbol{\theta}) = \sum_{\phi \in \Phi} \|\mathbf{A}_\phi \boldsymbol{\theta}_\phi\|_2.$$

When $\mathbf{A}_\phi$ is the identity operator, the penalty function $s$ is the overlapping group-lasso and $m = \text{Card}(\phi)$. When it is a discrete derivative operator, $s$ is a total variation penalty, and $m$ can be seen as the number of neighborhood relationships.

The non-smooth $\ell_{1,2}$-norm penalty can be approximated by a smooth function with known gradient computed using Nesterov's smoothing (Nesterov 2005b). Given a smoothness parameter $\mu > 0$, let us define the smooth approximation

$$s_\mu(\boldsymbol{\theta}) = \max_{\boldsymbol{\alpha} \in \mathcal{K}} \left\{ \boldsymbol{\alpha}^T \mathbf{A} \boldsymbol{\theta} - \frac{\mu}{2} \|\boldsymbol{\alpha}\|_2^2 \right\},$$

where $\mathcal{K}$ is the cartesian product of $\ell_2$-unit balls, $\mathbf{A}$ is the vertical concatenation of the matrices $\mathbf{A}_\phi$ and $\boldsymbol{\alpha}$ is an auxiliary variable resulting from the dual reformulation of $s(\boldsymbol{\theta})$. Note that $\lim_{\mu \to 0} s_\mu(\boldsymbol{\theta}) = s(\boldsymbol{\theta})$. A Fast Iterative Shrinkage-Thresholding Algorithm (FISTA, Beck and Teboulle 2009) step can then be applied after computing the gradient of the smooth part i.e. $g(\boldsymbol{\theta}) + \lambda_2 s_\mu(\boldsymbol{\theta})$ of the approximated criterion.

The main ingredient of CONESTA remains in the determination of the optimal smoothness parameter using the duality gap, which minimizes the number of FISTA iterations for a given precision $\epsilon$. The specification of $\mu$ is subject to dynamic update. A sequence of decreasing optimal smoothness parameters is generated in order to dynamically adapt the FISTA algorithm stepsize towards $\epsilon$. Namely, $\mu^k = \mu_{opt}(\epsilon^k)$. The smoothness parameter decreases as one gets closer to $\boldsymbol{\theta}^\star$, the solution of the problem defined in Equation 5. Since $\boldsymbol{\theta}^\star$ is unknown; the approximation of the distance to the minimum is achieved via the duality gap. Indeed

$$\mathrm{GAP}_{\mu^k}(\boldsymbol{\theta}^k) \geq f_{\mu^k}(\boldsymbol{\theta}^k) - f(\boldsymbol{\theta}^\star) \geq 0.$$

We refer the reader to the seminal paper for more details on the formulation of $\mathrm{GAP}_{\mu^k}(\boldsymbol{\theta}^k)$. The CONESTA routine is spelled out in the algorithm CONESTA solver where $L(g + \lambda_2 s_\mu)$ is the Lipschitz constant of $\nabla(g + \lambda_2 s_\mu)$, $k$ is the iteration counter for the inner FISTA updates and $i$ is the iteration counter for CONESTA updates.

### 3.2 Reformulation of MGLasso for CONESTA algorithm

Using CONESTA for solving the MGLasso problem requires a reformulation in order to comply with the form of loss function required by CONESTA. The objective of MGLasso can be written as

$$\operatorname{argmin} \frac{1}{2} ||\mathbf{Y} - \tilde{\mathbf{X}}\tilde{\boldsymbol{\beta}}||_2^2 + \lambda_1 ||\tilde{\boldsymbol{\beta}}||_1 + \lambda_2 \sum_{i<j} ||\boldsymbol{D}_{ij}\tilde{\boldsymbol{\beta}}||_2, \quad (6)$$

where $\mathbf{Y} = \mathrm{Vec}(\mathbf{X}) \in \mathbb{R}^{np}, \tilde{\boldsymbol{\beta}} = \mathrm{Vec}(\boldsymbol{\beta}) \in \mathbb{R}^{p(p-1)}, \tilde{\mathbf{X}}$ is a $\mathbb{R}^{[np] \times [p \times (p-1)]}$ block-diagonal matrix with $\mathbf{X}^{\setminus i}$ on the $i$-th block. The matrix $\boldsymbol{D}_{ij}$ is a $(p-1) \times p(p-1)$ matrix chosen so that $\boldsymbol{D}_{ij}\tilde{\boldsymbol{\beta}} = \boldsymbol{\beta}^i - \tau_{ij}\boldsymbol{\beta}^j$.

Note that we introduce this notation for simplicity of exposition, but, in practice, the sparsity of the matrices $\boldsymbol{D}_{ij}$ allows a more efficient implementation. Based on reformulation Equation 6, we may apply CONESTA to solve the objective of MGLasso for fixed $\lambda_1$ and $\lambda_2$. The procedure is applied, for fixed $\lambda_1$, to a range of decreasing values of $\lambda_2$ to obtain a hierarchical clustering. The corresponding pseudo-code is given in the following algorithm where $(\mathbf{X}^i)^\dagger$ denotes the pseudo-inverse of $\mathbf{X}^i$ and $\epsilon_{fuse}$ the threshold for merging clusters. We note here that problem in Equation 6 is of the same form as the optimization problem solved in the paper by Hadj-Selem et al. (2018): as they showed, CONESTA outperforms other optimization approaches such as the alternating direction method of multipliers (ADMM, Boyd et al. 2011), the excessive gap method (EGM, Nesterov 2005a), the classical FISTA with fixed smoothing and the inexact FISTA (Schmidt, Roux, and Bach 2011). Rather than repeating their experiments, we refer the reader to Section IV of their paper.

**Algorithm 1** CONESTA solver

**Inputs**:
  functions $g(\boldsymbol{\theta}), h(\boldsymbol{\theta}), s(\boldsymbol{\theta})$
  precision $\epsilon$
  penalty parameters $\lambda_1, \lambda_2$
  decreasing factor $\boldsymbol{\tau} \in (0, 1)$ for sequence of precisions
**Output:**
  $\boldsymbol{\theta}^{i+1} \in \mathbb{R}^d$
**Initializations:**
  $\boldsymbol{\theta}^0 \in \mathbb{R}^d$
  $\epsilon^0 = \boldsymbol{\tau} \text{GAP}_{\mu=10^{-8}}(\boldsymbol{\theta}^0)$
  $\mu^0 = \mu_{opt}(\epsilon^0)$
**repeat**
  $\epsilon_\mu^i = \epsilon^i - \mu^i \lambda_2 \frac{d}{2}$
  ▷ FISTA
  $k = 2$ ▷ new iterator
  $\boldsymbol{\theta}_{\text{FISTA}}^1 = \boldsymbol{\theta}_{\text{FISTA}}^0 = \boldsymbol{\theta}^i$ ▷ Initial parameters value
  $t_\mu = \frac{1}{L(g+\lambda_2 s_\mu)}$ ▷ Compute stepsize with $L(g + \lambda_2 s_\mu)$ the Lipschitz constant of $\nabla(g + \lambda_2 s_\mu)$
  **repeat**
    $\boldsymbol{z} = \boldsymbol{\theta}_{\text{FISTA}}^{k-1} + \frac{k-2}{k+1}(\boldsymbol{\theta}_{\text{FISTA}}^{k-1} - \boldsymbol{\theta}_{\text{FISTA}}^{k-2})$
    $\boldsymbol{\theta}_{\text{FISTA}}^k = \text{prox}_{\lambda_1 h}(\boldsymbol{z} - t_\mu \nabla(g + \lambda_2 s_\mu)(\boldsymbol{z}))$
  **until** $\text{GAP}_\mu(\boldsymbol{\theta}_{\text{FISTA}}^k) \leq \epsilon_\mu^i$
  $\boldsymbol{\theta}^{i+1} = \boldsymbol{\theta}_{\text{FISTA}}^k$
  $\epsilon^i = \text{GAP}_{\mu=\mu_i} \boldsymbol{\theta}^{i+1} + \mu^i \lambda_2 \frac{d}{2}$
  $\epsilon^{i+1} = \boldsymbol{\tau}\epsilon^i$
  $\mu^{i+1} = \mu_{opt}(\epsilon^{i+1})$
**until** $\epsilon^i \leq \epsilon$

## 3.3 Model selection

A crucial question for practical applications is the definition of a rule to select the penalty parameters $(\lambda_1, \lambda_2)$. This selection problem operates at two levels: $\lambda_1$ controls the sparsity of the graphical model, and $\lambda_2$ controls the number of clusters in the optimal clustering partition. These two parameters are dealt with separately: the sparsity parameter $\lambda_1$ is chosen via model selection, while the clustering parameter $\lambda_2$ varies across a grid of values in order to obtain graphs with different levels of granularity. The problem of model selection in graphical models is difficult in the high dimensional case where the number of samples is small compared to the number of variables, as classical Akaike information criterion (AIC, Akaike 1998) and Bayesian information criterion (BIC, Schwarz 1978) tend to perform poorly (Liu, Roeder, and Wasserman 2010).

In this paper, we focus on the StARS stability selection approach proposed by Liu, Roeder, and Wasserman (2010) as suggested by some preliminary tests where we compared the Extended BIC (EBIC, Foygel and Drton 2010), a model selection criterion calibrated with slope heuristics (Baudry, Maugis, and Michel 2012), the Rotation invariant criterion implemented in the Huge package (Zhao et al. 2012), the GGMSelect procedure (Giraud, Huet, and Verzelen 2012), cross-validation (Bien and Tibshirani 2011) and StARS. The method uses $k$ subsamples of data to estimate the associated graphs

**Algorithm 2** MGLasso algorithm

**Inputs**:
  Set of variables $\mathbf{X} = \{\mathbf{X}^1, \dots, \mathbf{X}^p\} \in \mathbb{R}^{n \times p}$
  Penalty parameters $\lambda_1 \geq 0, \lambda_{2\text{initial}} > 0$
  Increasing factor $\eta > 1$ for fusion penalties $\lambda_2$
  Fusion threshold $\epsilon_{fuse} \geq 0$
**Outputs:** For $\lambda_1$ fixed and $\lambda_2$ from 0 to $\lambda_{2\text{initial}} \times \eta^{(I)}$ with $I$ the number of iterations:
  Regression vectors $\boldsymbol{\beta}(\lambda_1, \lambda_2) \in \mathbb{R}^{p \times (p-1)}$,
  Clusters partition of variables indices in $K$ clusters: $C(\lambda_1, \lambda_2)$
**Initializations:**
  $\boldsymbol{\beta}^i = (\mathbf{X}^i)^\dagger \mathbf{X}^i$, $\forall i = 1, \dots, p$ for warm start in CONESTA solver
  $C = \{\{1\}, \dots, \{p\}\}$ Initial clusters with one element per cluster.
  Set $\lambda_2 = 0$
  Compute $\boldsymbol{\beta}$ using CONESTA solver
  Update clusters $C$ with rule described in **while** loop.
**Set:** $\lambda_2 = \lambda_{2\text{initial}}$
  ▷ Clustering path
**while** Card$(C) > 1$ **do**
  Compute $\boldsymbol{\beta}$ using CONESTA solver with warm start from previous iteration
  ▷ Clusters update
  Compute pairwises distances $d(i,j) = \left\| \boldsymbol{\beta}^i - \boldsymbol{\tau}_{ij} \boldsymbol{\beta}^j \right\|_2$, $\forall i, j \in \{1, \dots, p\}$
  Determine clusters $C_k (k = 1, \dots, K)$ with the rule $(i,j) \in C_k$ iff. $d(i,j) \leq \epsilon_{fuse}$
  $\lambda_2 = \lambda_2 \times \nu$
**end while**

for a given range of $\lambda_1$ values. For each value, a global instability of the graph edges is computed. The optimal value of $\lambda_1$ is chosen so as to minimize the instability, as follows. Let $\lambda_1^{(1)}, \dots, \lambda_1^{(K)}$ be a grid of sparsity regularization parameters, and $S_1, \dots, S_N$ be the $N$ bootstrap samples obtained by sampling the rows of the data set $\mathbf{X}$. For each $k \in \{1, \dots, K\}$ and for each $j \in \{1, \dots, N\}$, we denote by $\mathcal{A}^{k,j}(\mathbf{X})$ the adjacency matrix of the estimated graph obtained by applying the inference algorithm to $S_n$ with regularization parameter $\lambda_1^{(k)}$. For each possible edge $(s,t) \in \{1, \dots, p\}^2$, the probability of edge appearance is estimated empirically by

$$\hat{\theta}_{st}^{(k)} = \frac{1}{N} \sum_{j=1}^{N} \mathcal{A}_{st}^{k,j}.$$

Define

$$\hat{\xi}_{st}(\Lambda) = 2\hat{\theta}_{st}(\Lambda) \left(1 - \hat{\theta}_{st}(\Lambda)\right)$$

$$\hat{\xi}_{st}(\lambda_1^{(k)}) = 2\hat{\theta}_{st}^{(k)} \left(1 - \hat{\theta}_{st}^{(k)}\right)$$

the empirical instability of edge $(s,t)$ (that is, twice the variance of the Bernoulli indicator of edge $(s,t)$). The instability level associated with $\lambda_1^{(k)}$ is given by

$$\hat{D}(\lambda_1^{(k)}) = \frac{\sum_{s<t} \hat{\xi}_{st}(\lambda_1^{(k)})}{\binom{p}{2}}.$$

StARS selects the optimal penalty parameter as follows

$$\hat{\lambda} = \max_k \left\{ \lambda_1^{(k)} \; : \; \hat{D}(\lambda_1^{(k)}) \leq \upsilon, k \in \{1, \dots, K\} \right\},$$

where $\upsilon$ is the threshold chosen for the instability level.

# 4 Simulation experiments

In this Section, we conduct a simulation study to evaluate the performance of the MGLasso method, both in terms of clustering and support recovery. Receiver Operating Characteristic (ROC) curves are used to evaluate the adequacy of the inferred graphs with the ground truth for the MGLasso and GLasso in its neighborhood selection version in the Erdös-Rényi (Erdős, Rényi, et al. 1960), Scale-free (Newman, Strogatz, and Watts 2001), and Stochastic Block Models (SBM, Fienberg and Wasserman 1981) frameworks. The Adjusted Rand indices are used to compare the partitions obtained with MGLasso, hierarchical agglomerative clustering, and K-means clustering in a stochastic block model framework.

## 4.1 Synthetic data models

We consider three different synthetic network models: the Stochastic Block Model (Fienberg and Wasserman 1981), the Erdös-Renyi model (Erdős, Rényi, et al. 1960) and the Scale-Free model (Newman, Strogatz, and Watts 2001). In each case, Gaussian data is generated by drawing $n$ independent realizations of a multivariate Gaussian distribution $\mathcal{N}(0, )$ where $\in \mathbb{R}^{p \times p}$ and $= ^{-1}$. The support of , equivalent to the network adjacency matrix, is generated from the three different models. The difficulty level of the problem is controlled by varying the ratio $\frac{n}{p}$ with $p$ fixed at 40: $\frac{n}{p} \in \{0.5, 1, 2\}$.

### 4.1.1 Stochastic Block Model

We construct a block-diagonal precision matrix  as follows. First, we generate the support of  as shown in Figure 2, denoted by $\boldsymbol{A} \in \{0, 1\}^{p \times p}$. To do this, the variables are first partitioned into $K = 5$ hidden groups, noted $C_1, \dots, C_K$ described by a latent random variable $Z_i$, such that $Z_i = k$ if $i = C_k$. $Z_i$ follows a multinomial distribution

$$P(Z_i = k) = \pi_k, \quad \forall k \in \{1, \dots, K\},$$

where $\pi = (\pi_1, \dots, \pi_k)$ is the vector of proportions of clusters whose sum is equal to one. The set of latent variables is noted $\mathbf{Z} = \{Z_1, \dots, Z_K\}$. Conditionally to $\mathbf{Z}$, $A_{ij}$ follows a Bernoulli distribution such that

$$A_{ij} | Z_i = k, Z_j = l \sim \mathcal{B}(\alpha_{kl}), \quad \forall k, l \in \{1, \dots, K\},$$

where $\alpha_{kl}$ is the probability of inter-cluster connectivity, with $\alpha_{kl} = 0.01$ if $k \neq l$ and $\alpha_{ll} = 0,75$. For $k \in \{1, \dots, K\}$, we define $p_k = \sum_{i=1}^{p} \mathbf{1}_{\{Z_i = k\}}$. The precision matrix  of the graph is then calculated as follows. We define $\Omega_{ij} = 0$ if $Z_i \neq Z_j$ ; otherwise, we define $\Omega_{ij} = A_{ij}\omega_{ij}$ where, for all $i \in \{1, \dots, p\}$ and for all $j \in \{1, \dots, p | Z_j = Z_i\}$, $\omega_{ij}$ is given by :

$$\begin{aligned} \omega_{ii} &:= \frac{1 + \rho(p_{Z_i} - 2)}{1 + \rho(p_{Z_i} - 2) - \rho^2(p_{Z_i} - 1)}; \\ \omega_{ij} &:= \frac{-\rho}{1 + \rho(p_{Z_i} - 2) - \rho^2(p_{Z_i} - 1)}. \end{aligned}$$

If $\alpha_{ll}$ were to be equal to one, this construction of  would make it possible to control the level of correlation between the variables in each block to $\rho$. Introducing a more realistic scheme with $\alpha_{ll} = 0.75$ allows only to have an approximate control.

```
library(mglasso)
set.seed(2020)
sim_sbm <- sim_data(
  p = 40,
  structure = "block_diagonal",
  alpha = rep(1 / 5, 5),
  prob_mat = diag(0.75, 5),
  rho = 0.2,
  inter_cluster_edge_prob = 0.01
)
gsbm <- adj_mat(sim_sbm$graph)
Matrix::image(
  as(gsbm, "sparseMatrix"),
  sub = "",
  xlab = "",
  ylab = ""
)
```

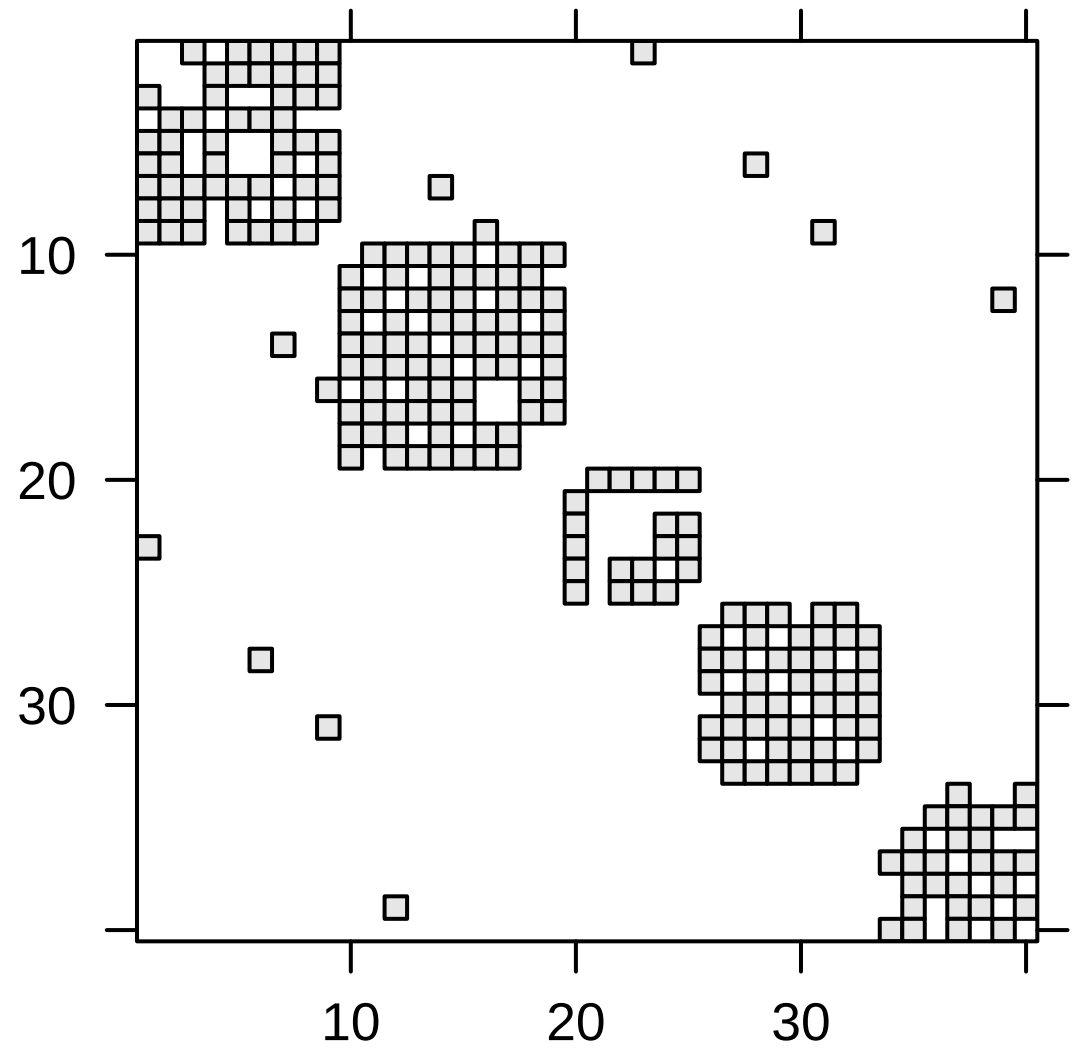

Figure 2: Adjacency matrix of a stochastic block model defined by $K = 5$ classes with identical prior probabilities set to $\pi = 1/K$, inter-classes connection probability $\alpha_{kl} = 0.01, k \neq l$, intra-classes connection probability $\alpha_{ll} = 0.75$ and $p = 40$ vertices.

### 4.1.2 Erdös-Renyi Model

The Erdös-Renyi model is a special case of the stochastic block model where $\alpha_{kl} = \alpha_{ll} = \alpha$ is constant. We set the density $\alpha$ of the graph to 0.1; see Figure 3 for an example of the graph resulting from this model.

```
set.seed(2022)
sim_erdos <- sim_data(p = 40, structure = "erdos", p_erdos = 0.1)
gerdos <- adj_mat(sim_erdos$graph)
```

```
Matrix::image(as(gerdos, "sparseMatrix"), sub = "", xlab = "", ylab = "")
```

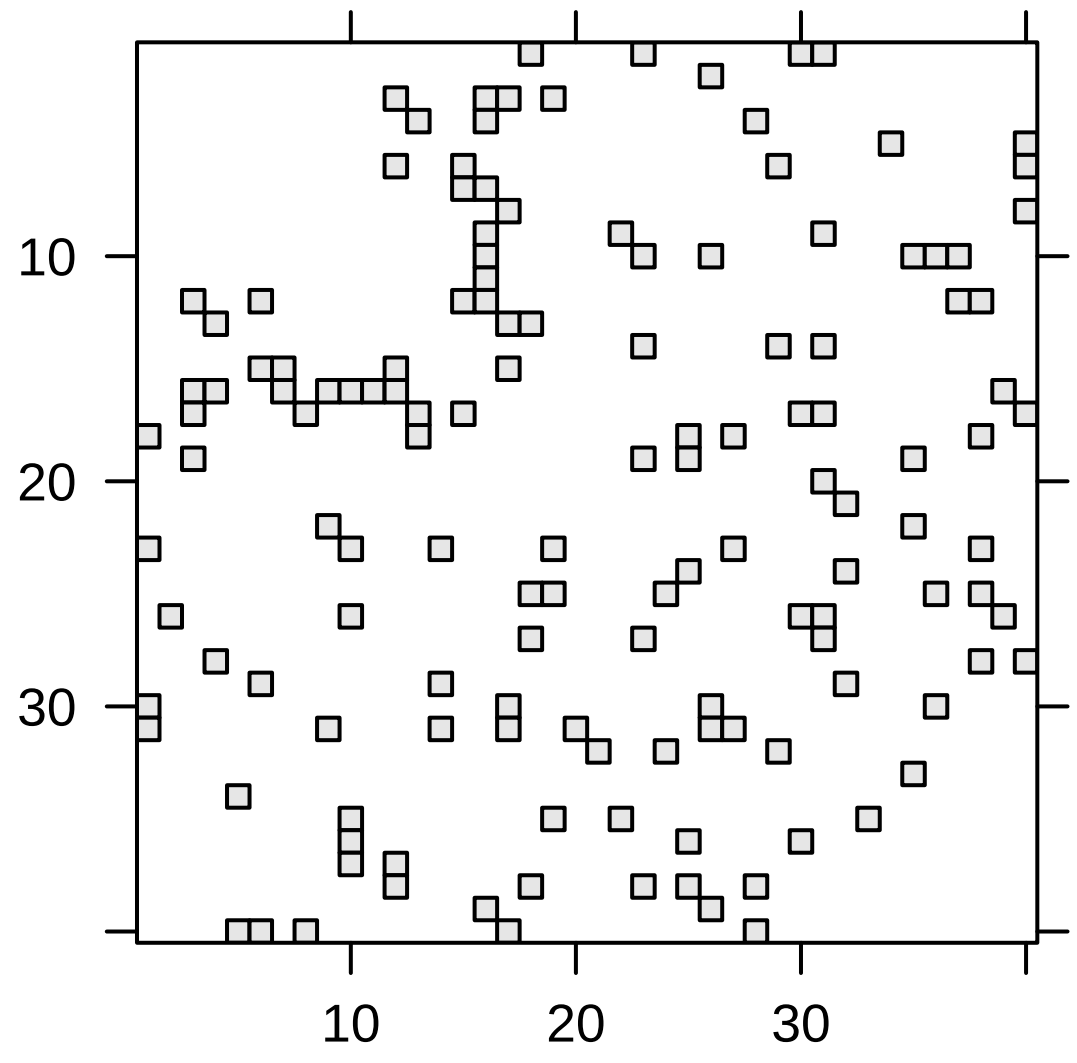

Figure 3: Adjacency matrix of an Erdös-Renyi model with probability of connection $\alpha = 0.1$ and $p = 40$ vertices.

### 4.1.3 Scale-free Model

The Scale-free Model generates networks whose degree distributions follow a power law. The graph starts with an initial chain graph of 2 nodes. Then, new nodes are added to the graph one by one. Each new node is connected to an existing node with a probability proportional to the degree of the existing node. We set the number of edges in the graph to 40. An example of scale-free graph is shown in Figure 4.

```
set.seed(2022)
sim_sfree <- sim_data(p = 40, structure = "scale_free")
gsfree <- adj_mat(sim_sfree$graph)
Matrix::image(as(gsfree, "sparseMatrix"), sub = "", xlab = "", ylab = "")
```

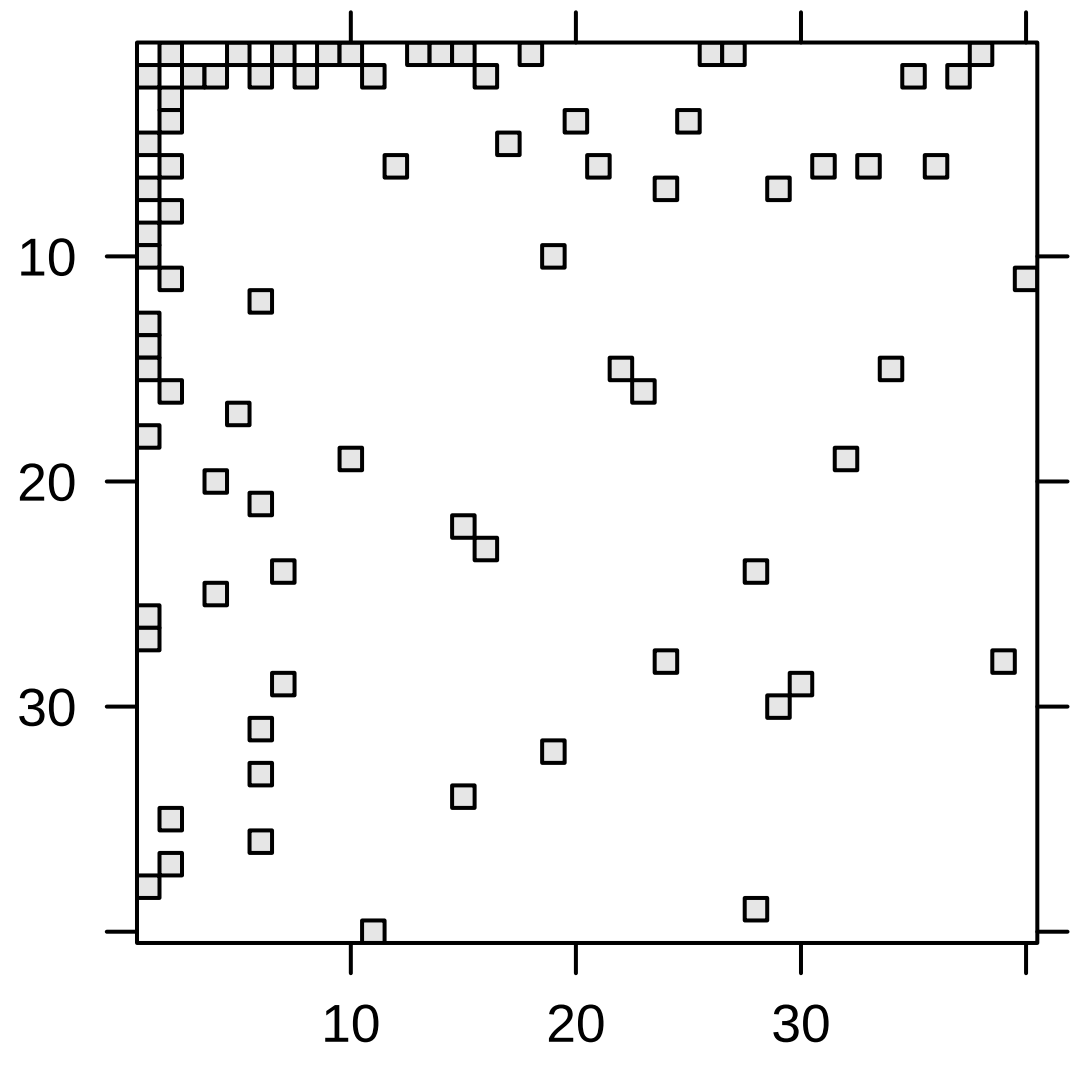

Figure 4: Adjacency matrix of a Scale-free model with 40 edges and $p = 40$ nodes.

## 4.2 Support recovery

We compare the network structure learning performance of our approach to that of GLasso in its neighborhood selection version using ROC curves. In both GLasso and MGLasso, the sparsity is controlled by a regularization parameter $\lambda_1$; however, MGLasso admits an additional regularization parameter, $\lambda_2$, which controls the strength of convex clustering. To compare the two methods, in each ROC curve, we vary the parameter $\lambda_1$ while the parameter $\lambda_2$ (for MGLasso) is kept constant. We computed ROC curves for 4 different penalty levels for the $\lambda_2$ parameter; since GLasso does not depend on $\lambda_2$, the GLasso ROC curves are replicated.

In a decision rule associated with a sparsity penalty level $\lambda_1$, we recall the definition of the two following functions. The true positive rate is given by $\frac{TP(\lambda_1)}{TP(\lambda_1)+FN(\lambda_1)}$. The false positive rate is defined as follows $1-\frac{TN(\lambda_1)}{TN(\lambda_1)+FP(\lambda_1)}$, where $TP$ is the number of true positives, $TN$ the number of true negatives, $FN$ the number of false negatives and $FP$ the number of false positives. The ROC curve represents the true positive rate as a function of the false positive rate. For a given level of true positive rate, the best method minimizes the false positive rate.

For each configuration ($n, p$ fixed), we generate 50 replications and their associated ROC curves, which are then averaged. The average ROC curves for the three models are given in Figure 5, Figure 6 and Figure 7 by varying $\frac{n}{p} \in \{0.5, 1, 2\}$.

```
library(ggplot2)
library(ghibli)
load("./data/roc_dtf_erdos.RData")
np.labs <- c("frac(n, p) == 0.5", "frac(n, p) == 1", "frac(n, p) == 2")
names(np.labs) <- c("0.5", "1", "2")
tv.labs <- c("lambda[2] == 0", "lambda[2] == 3.33", "lambda[2] == 10")
names(tv.labs) <- c("0", "3.33", "10")
roc_dtf_erdos <- dplyr::filter(roc_dtf_erdos, tv != 6.67)
```

```r
ggplot(roc_dtf_erdos, aes(
  x     = 100 * fpr,
  y     = 100 * tpr,
  color = method
)) +
  geom_line(linewidth = 0.7) +
  facet_grid(np ~ tv, labeller = labeller(
    np = as_labeller(np.labs, label_parsed),
    tv = as_labeller(tv.labs, label_parsed)
  )) +
  geom_abline(
    intercept = 0,
    slope = 1,
    linetype = "dashed",
    color = "grey"
  ) +
  xlab("False Positive Rate") +
  ylab("True Positive Rate") +
  ggtitle("") +
  scale_colour_manual(
    name = "Method",
    labels = c("GLasso", "MGLasso"),
    values = ghibli::ghibli_palette("MarnieMedium1")[5:6]
  )
```

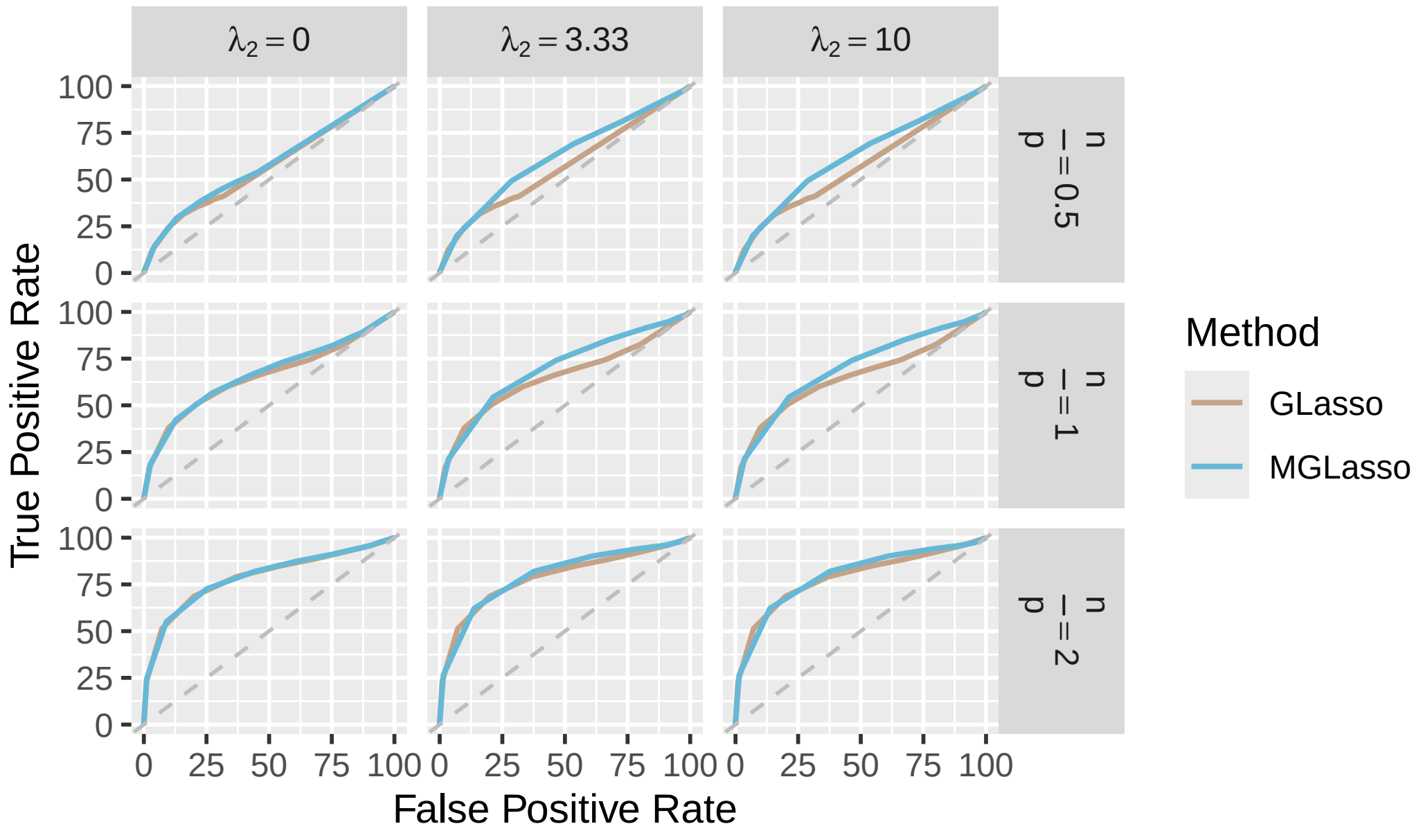

Figure 5: Mean ROC curves for MGLasso and GLasso graph inference in the Erdös-Renyi model. We varied the fusion penalty parameter of MGLasso $\lambda_2 \in \{0, 3.33, 10\}$ alongside the ratio $\frac{n}{p} \in \{0.5, 1, 2\}$. Within each panel, the ROC curve shows the True positive rate (y-axis) vs. the False positive rate (x-axis) for both MGLasso (blue) and GLasso (brown). Since GLasso does not have a fusion penalty, its ROC curves were replicated for panels belonging to the same row. We also plot the random classifier (dotted grey line). The results have been averaged over 50 simulated datasets and suggest that MGLasso performs no worse than GLasso. For $\lambda_2 = 0$, the MGLasso approach is equivalent to GLasso in its neighborhood selection version.

```
load("./data/roc_dtf_sfree.RData")
np.labs <- c("frac(n, p) == 0.5", "frac(n, p) == 1", "frac(n, p) == 2")
names(np.labs) <- c("0.5", "1", "2")
tv.labs <- c("lambda[2] == 0", "lambda[2] == 3.33", "lambda[2] == 10")
names(tv.labs) <- c("0", "3.33", "10")
roc_dtf_sfree <- dplyr::filter(roc_dtf_sfree, tv != 6.67)
ggplot(roc_dtf_sfree, aes(
  x     = 100 * fpr,
  y     = 100 * tpr,
  color = method
)) +
  geom_line() +
  facet_grid(np ~ tv, labeller = labeller(
    np = as_labeller(np.labs, label_parsed),
    tv = as_labeller(tv.labs, label_parsed)
  )) +
  geom_abline(
    intercept = 0,
    slope = 1,
    linetype = "dashed",
    color = "grey"
```

```
  ) +
  xlab("False Positive Rate") +
  ylab("True Positive Rate") +
  ggtitle("") +
  scale_colour_manual(
    name = "Method",
    labels = c("GLasso", "MGLasso"),
    values = ghibli_palette("MarnieMedium1")[5:6]
  )
```

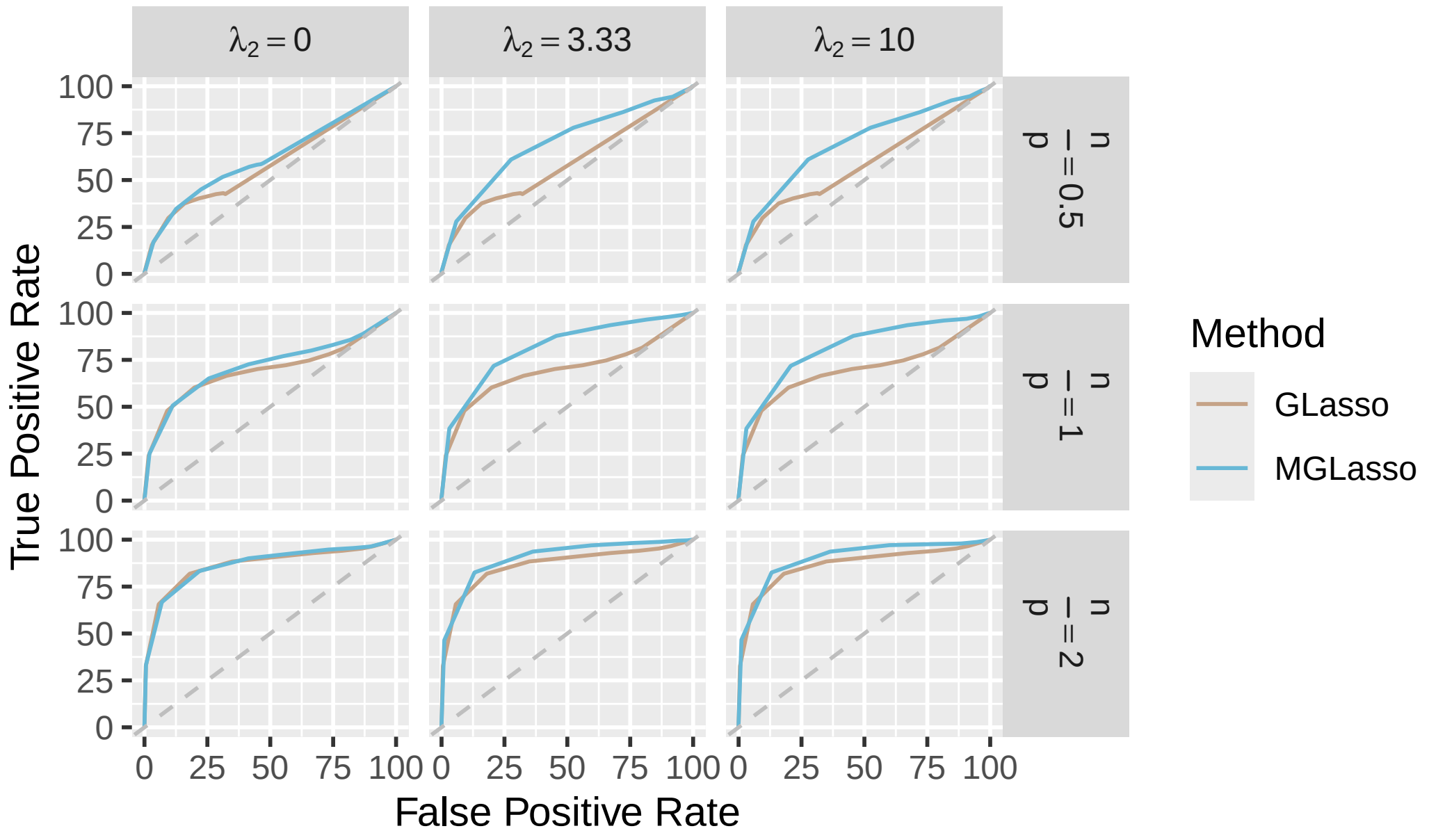

Figure 6: Mean ROC curves for MGLasso and GLasso graph inference in the Scale-free model. We varied the fusion penalty parameter of MGLasso $\lambda_2 \in \{0, 3.33, 10\}$ alongside the ratio $\frac{n}{p} \in \{0.5, 1, 2\}$. Within each panel, the ROC curve shows the True positive rate (y-axis) vs. the False positive rate (x-axis) for both MGLasso (blue) and GLasso (brown). Since GLasso does not have a fusion penalty, its ROC curves were replicated for panels belonging to the same row. We also plot the random classifier (dotted grey line). The results have been averaged over 50 simulated datasets and suggest that MGLasso performs no worse than GLasso. For $\lambda_2 = 0$, the MGLasso approach is equivalent to Glasso in its neighborhood selection version.

```
load("./data/roc_dtf_sbm.RData")
np.labs <- c("frac(n, p) == 0.5", "frac(n, p) == 1", "frac(n, p) == 2")
names(np.labs) <- c("0.5", "1", "2")
tv.labs <- c("lambda[2] == 0", "lambda[2] == 3.33", "lambda[2] == 10")
names(tv.labs) <- c("0", "3.33", "10")
roc_dtf_sbm <- dplyr::filter(roc_dtf_sbm, tv != 6.67)
ggplot(roc_dtf_sbm, aes(
  x     = 100 * fpr,
  y     = 100 * tpr,
  color = method
)) +
  geom_line() +
```

```
facet_grid(np ~ tv, labeller = labeller(
  np = as_labeller(np.labs, label_parsed),
  tv = as_labeller(tv.labs, label_parsed)
)) +
geom_abline(
  intercept = 0,
  slope = 1,
  linetype = "dashed",
  color = "grey"
) +
xlab("False Positive Rate") +
ylab("True Positive Rate") +
ggtitle("") +
scale_colour_manual(
  name = "Method",
  labels = c("GLasso", "MGLasso"),
  values = ghibli_palette("MarnieMedium1")[5:6]
)
```

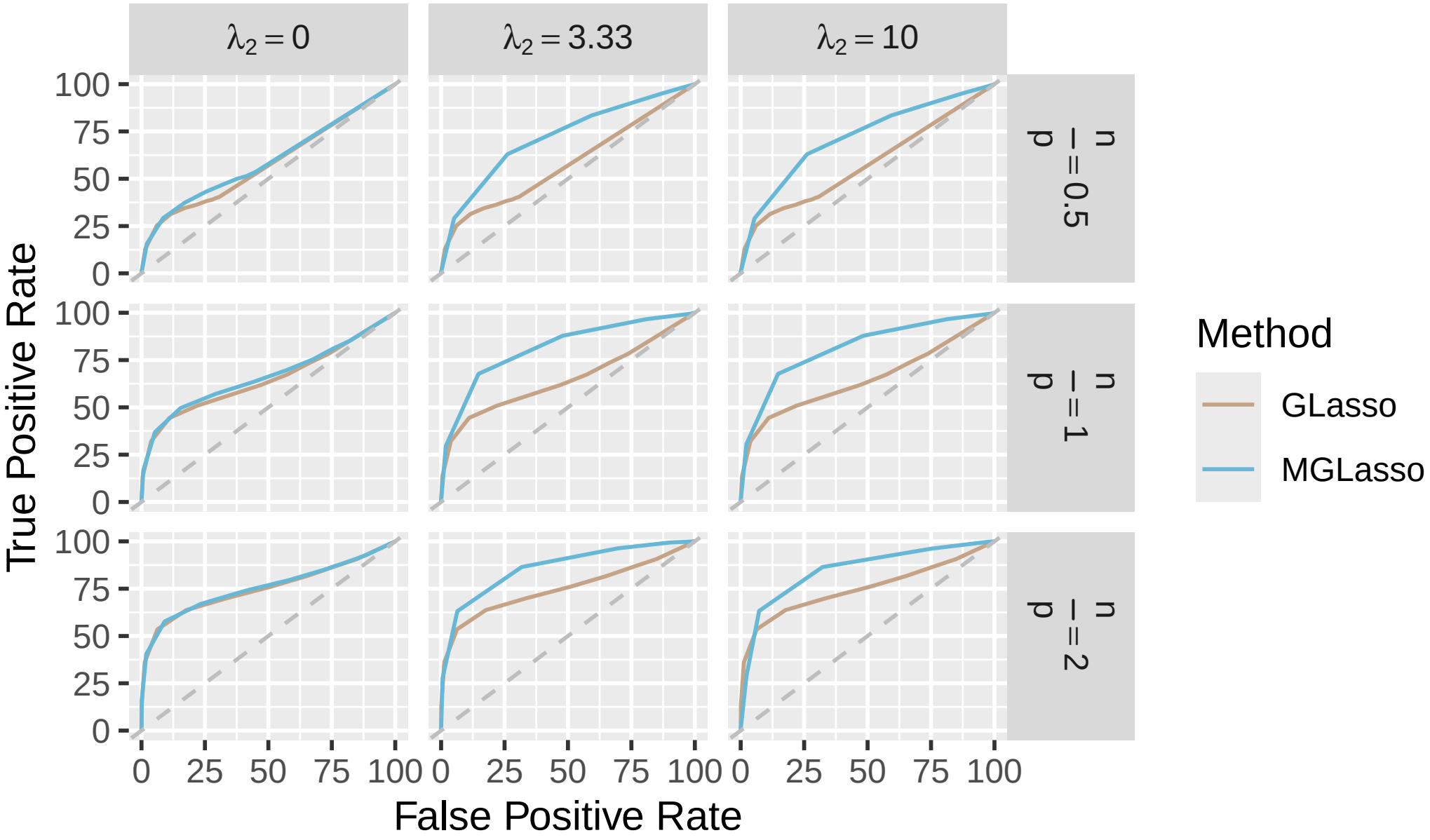

Figure 7: Mean ROC curves for MGLasso and GLasso graph inference in the stochastic block model. We varied the fusion penalty parameter of MGLasso $\lambda_2 \in \{0, 3.33, 10\}$ alongside the ratio $\frac{n}{p} \in \{0.5, 1, 2\}$. Within each panel, the ROC curve shows the True positive rate (y-axis) vs. the False positive rate (x-axis) for both MGLasso (blue) and GLasso (brown). Since GLasso does not have a fusion penalty, its ROC curves were replicated for panels belonging to the same row. We also plot the random classifier (dotted grey line). The results have been averaged over 50 simulated datasets and suggest that MGLasso performs no worse than GLasso. For $\lambda_2 = 0$, the MGLasso approach is equivalent to Glasso in its neighborhood selection version.

Based on these empirical results, we first observe that, in all the considered simulation models, MGLasso improves over GLasso in terms of support recovery in the high-dimensional setting where $p < n$. In addition, in the absence of a fusion penalty, i.e., $\lambda_2 = 0$, MGLasso performs no worse than GLasso in each of the 3 models. However, for $\lambda_2 > 0$, increasing penalty value does not seem to significantly improve the support recovery performances for the MGLasso, as we observe similar results for $\lambda_2 = 3.3, 10$. Preliminary analyses show that, as $\lambda_2$ increases, the estimates of the regression vectors are shrunk towards 0. This shrinkage effect of group-fused penalty terms was also observed in (Chu et al. 2021). Note that the performance of the MGLasso deteriorates comparatively to GLasso when the inter-clusters edge connection probability of the stochastic block model is high.

## 4.3 Clustering

In order to study clustering performance, we compared the partitions estimated by MGLasso, Hierarchical Agglomerative Clustering (HAC) with Ward's distance and K-means to the true partition in a stochastic block model framework. Euclidean distances between variables are used for HAC and K-means. The criterion used for the comparison is the adjusted Rand index (ARI). We studied the influence of the correlation level inside clusters on the clustering performances through two different parameters: $\rho \in \{0.1, 0.3\}$; the vector of cluster proportions is fixed at $= (1/5, \dots, 1/5)$. Hundred Gaussian data sets were then simulated for each configuration ($\rho$, $n/p$ fixed).The optimal sparsity penalty for MGLasso was chosen by the Stability Approach to Regularization Selection (StARS) method (Liu, Roeder, and Wasserman 2010). In practice, we estimated a stability-like parameter in a sample of graphs simulated via the stochastic block model. This estimation of edge variability was then used as the threshold for the StARS method. The parameter $\lambda_2$ has been varied.

```
load("./data/rand_dt_lower_cor_sbm.RData")
plot_res(
  dt_rand,
  crit_ = "rand",
  ncluster_ = c(5, 10, 15, 20),
  cor_ = 0.25,
  np_ = c(0.5, 1, 2),
  main = ""
)
```

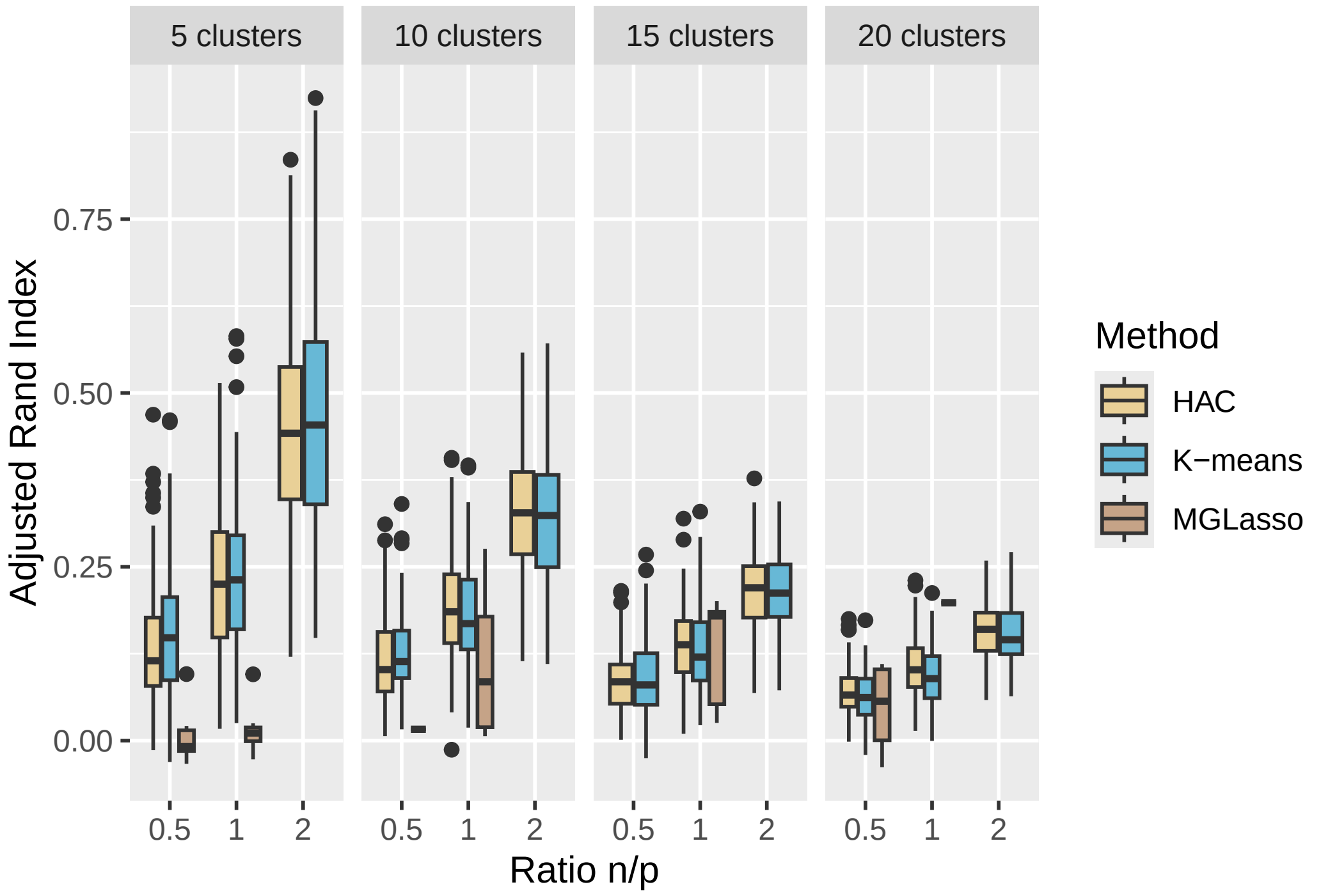

Figure 8: Boxplots of Adjusted Rand Indices for the stochastic block model with 5 classes and $p = 40$ variables for a correlation level $\rho = 0.1$. The number of estimated clusters $\{5, 10, 15, 20\}$ vary alongside the ratio $\frac{n}{p} \in \{0.5, 1, 2\}$. Within each panel, the boxplots of ARI between true partition (with 5 classes) and estimated clustering partitions on 100 simulated datasets for $k$-means (blue), hierarchical agglomerative clustering (yellow), and MGLasso (brown) methods are plotted against the ratio $\frac{n}{p}$. The cluster assignments of MGLasso are computed from a distance between estimated regression vectors for a given value of $\lambda_2$. Missing boxplots for MGLasso thus mean computed partitions in the grid of values of $\lambda_2$ do not yield the fixed number of clusters. The higher the ARI values, the better the estimated clustering partition is.

```
load("./data/rand_dt_higher_cor_sbm.RData")
plot_res(
  dt_rand,
  crit_ = "rand",
  ncluster_ = c(5, 10, 15, 20),
  cor_ = 0.95,
  np_ = c(0.5, 1, 2),
  main = ""
)
```

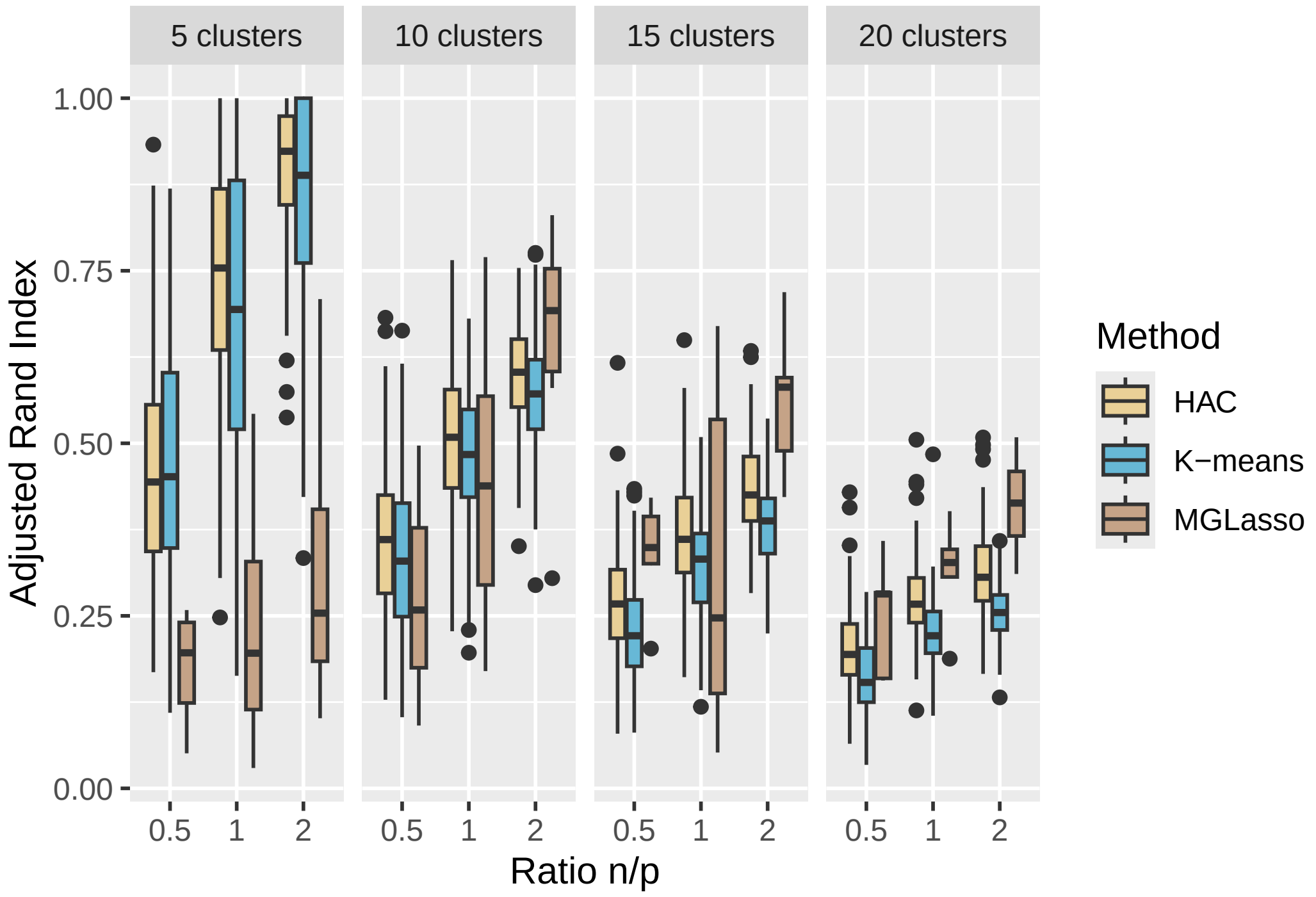

Figure 9: Boxplots of Adjusted Rand Indices for the stochastic block model with 5 classes and $p = 40$ variables for a correlation level $\rho = 0.3$. The number of estimated clusters $\{5, 10, 15, 20\}$ vary alongside the ratio $\frac{n}{p} \in \{0.5, 1, 2\}$. Within each panel, the boxplots of ARI between true partition (with 5 classes) and estimated clustering partitions on 100 simulated datasets for $k$-means (blue), hierarchical agglomerative clustering (yellow), and MGLasso (brown) methods are plotted against the ratio $\frac{n}{p}$. The cluster assignments of MGLasso are computed from a distance between estimated regression vectors for a given value of $\lambda_2$. The higher the ARI values, the better the estimated clustering partition is.

The expected empirical evidence that MGLasso would work reasonably well for strongly correlated variables is somehow highlighted in Figure 8 and Figure 9. The performances of MGLasso slightly improve when going from Figure 8 to Figure 9, which corresponds to correlation levels of 0.1 and 0.3 between variables belonging to the same block, respectively. We observe the same trend for the HAC and the k-means. Compared to these two approaches, the MGLasso presents the lowest values of adjusted Rand indices, thus suggesting a lower quality of clustering. It should be noted that the performance of MGLasso can be sensitive to the selection of the Lasso penalty parameter and the threshold fixed to determine clusters' fusion. In practice, this fusion threshold is varied in a grid of values close to zero and lower than $10^{-3}$. The value leading to the maximum number of intermediate clusters in the clustering path is chosen. Using non-trivial weights could also improve the overall performance of MGLasso.

During the revision of this paper, an interesting question was raised regarding the behavior of the algorithm in a phylogenetic-based model. To investigate this, extensive numerical experiments were conducted on a phylogenetic-based model that evaluates only clustering performances. The results showed that the MGLASSO algorithm's performance improves, and the method performs as well as some state-of-the-art clustering approaches, including vanilla convex clustering and spectral clustering. In phylogenetic-based models, adjusted Rand indices can be computed between the estimated partition with $k$ clusters and the true partition in $k$ clusters computed from the tree used for the simulation procedure. This differs from the clustering performance evaluation scheme applied in the stochastic block model, where the true partition is considered fixed.

# 5 Applications

To illustrate the proposed simultaneous graphs and clusters inference approach, we present analyses where the MGLasso model is applied to microbial association data for the study of multiscale networks between operational taxonomic units and to transcriptomic and methylation genotypes for multi-omics data integration.

## 5.1 Application to microbial associations in gut data

We analyze microbial associations in human gut microbiome data acquired from the round 1 of the American Gut Project (AGP, McDonald et al. (2018)) for $p = 127$ operational taxonomic units (OTUs) and $n = 289$ individuals samples. The count of microbial OTUs is an indicator of the abundance of underlying microbial populations. Here, we investigate the network and clustering structures of the OTUs for different levels of granularity on the processed data included in the SpiecEasi R package (see Kurtz (2015) for details). The data is first normalized to have a unit-sum per sample and to remove biases. Then, a centered log-ratio (clr, Aitchison 1982) transformation with an added unit pseudo-count is applied to come back to an unconstrained Euclidean space. For fitting the MGLasso model, we select the Lasso penalty parameter $\lambda_1$ via the StARS approach with threshold $v = 0.05$ and vary the fusion penalty $\lambda_2$ in the interval $[0, 20]$ with irregular steps. The CPU time taken for 20 values of $\lambda_2$ is about 8 hours with parallel evaluations on a computation cluster with as many cores as $\lambda_2$ values. The maximal number of iterations is set to 10000 and the solver precision to 0.01.

We finally illustrate our new method of inferring the multiscale Gaussian graphical model, with an application to the analysis of microbial associations in the American Gut Project. The data used are count data that have been previously normalized by applying the log-centered ratio technique as used in (Kurtz 2015). After some filtering steps (Kurtz 2015) on the operational taxonomic units (OTUs) counts (removed if present in less than 37% of the samples) and the samples (removed if sequencing depth below 2700), the top OTUs are grouped in a dataset composed of $n = 289$ for 127 OTUs. As a preliminary analysis, we perform a hierarchical agglomerative clustering (HAC) on the OTUs, which allows us to identify four significant groups. The correlation matrix of the dataset is given in fig-emp-cor; variables have been rearranged according to the HAC partition.

Using these settings, we compute a clustering path of the solutions and estimated graphs for 5 values of $\lambda_2$ corresponding to 5 different clusters partitions. The Figure 10 shows how the predicted $\hat{\boldsymbol{X}}$ evolves through $\lambda_2$. The $\hat{\boldsymbol{X}}$ are computed from estimated centroids $\hat{\boldsymbol{\beta}}$ and projected onto two principal components of the original data. The path is not always agglomerative, but the clusters' splits observed ensure optimal solutions.

The Figure 11 displays graphs and clusters for different levels of granularity: 127, 63, 31, 15 and 2 clusters. For computing the clusters' assignment of nodes, the fusion threshold has been set to $\epsilon_{fuse} = 0.001$. Variables that belong to the same cluster share the same neighborhood; thus, the neighboring information is summarized into a single variable representative of the group. The subfigures show graphs at multiple levels of granularity which are built on the meta-variables or representative variables.

To assess the relevance of the inferred clusters, they are compared to known taxonomic ranks (phylum, class, order, family, genera, or species). The phylum classification is used. For example, for a clustering partition in 2 groups, the MGLasso clustering partition is composed of 120 variables versus 7 variables. The cluster 2 is exclusively composed of OTUs belonging to the Proteobacteria

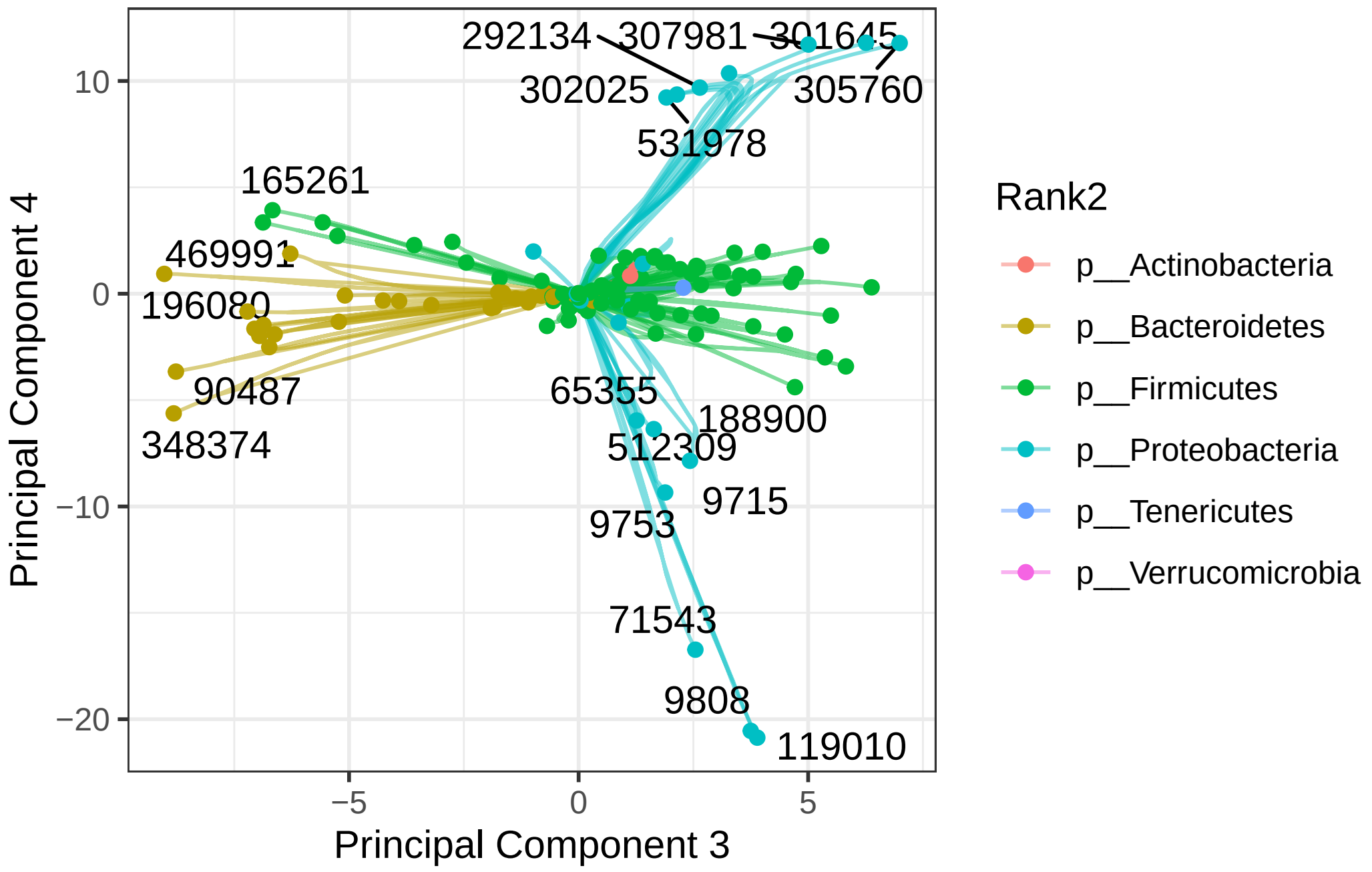

Figure 10: Clustering path of the MGLasso convex clustering solutions on microbiome data with 127 OTUs. The predicted data are projected onto the two principal components of the original data, while the fusion penalty varies. As $\lambda_2$ increases, it reaches a value for which all the estimated centroids are equal; thus, the branches of the path converge to a unique point in the center of the graph. OTUs are colored according to their phylum classification. The path displays abrupt merges. The pure cluster on the graph's left side (down) corresponds to the phylum Bacteroidetes.

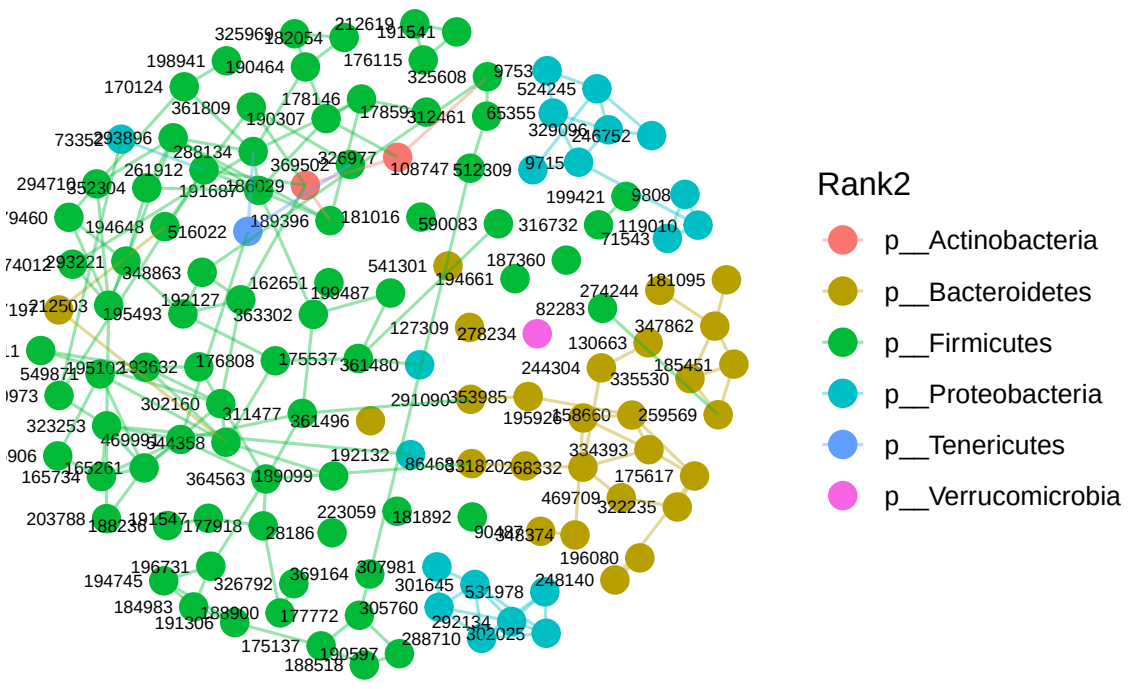

(a) 127 clusters graph

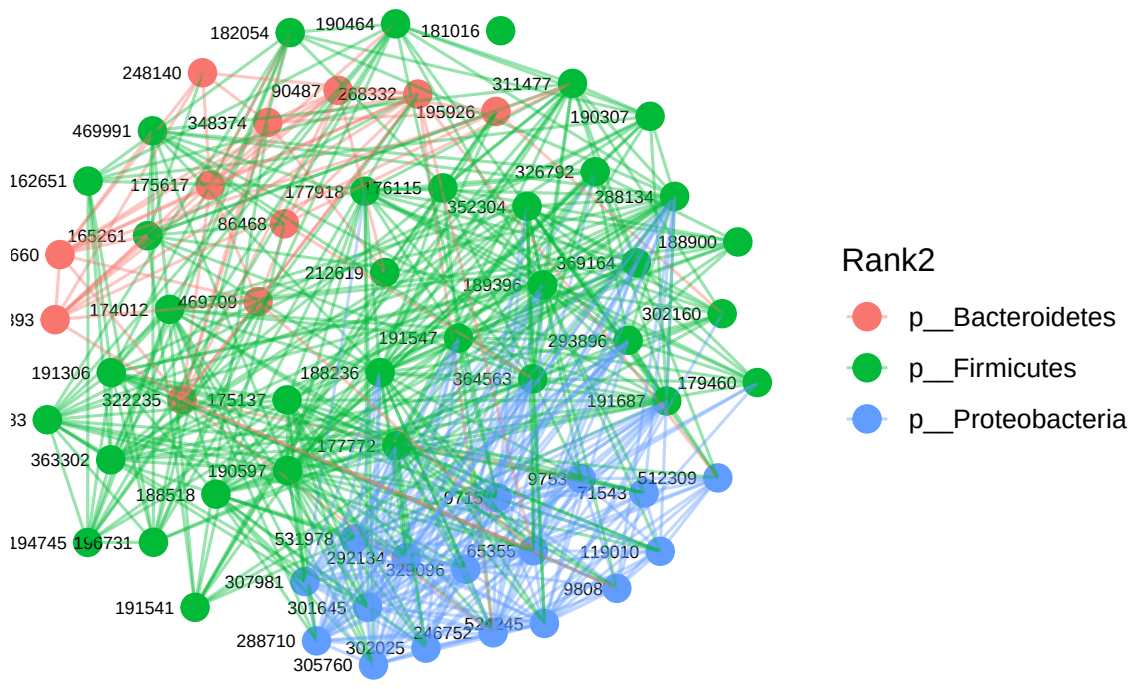

(b) Meta-variables graph with 63 clusters

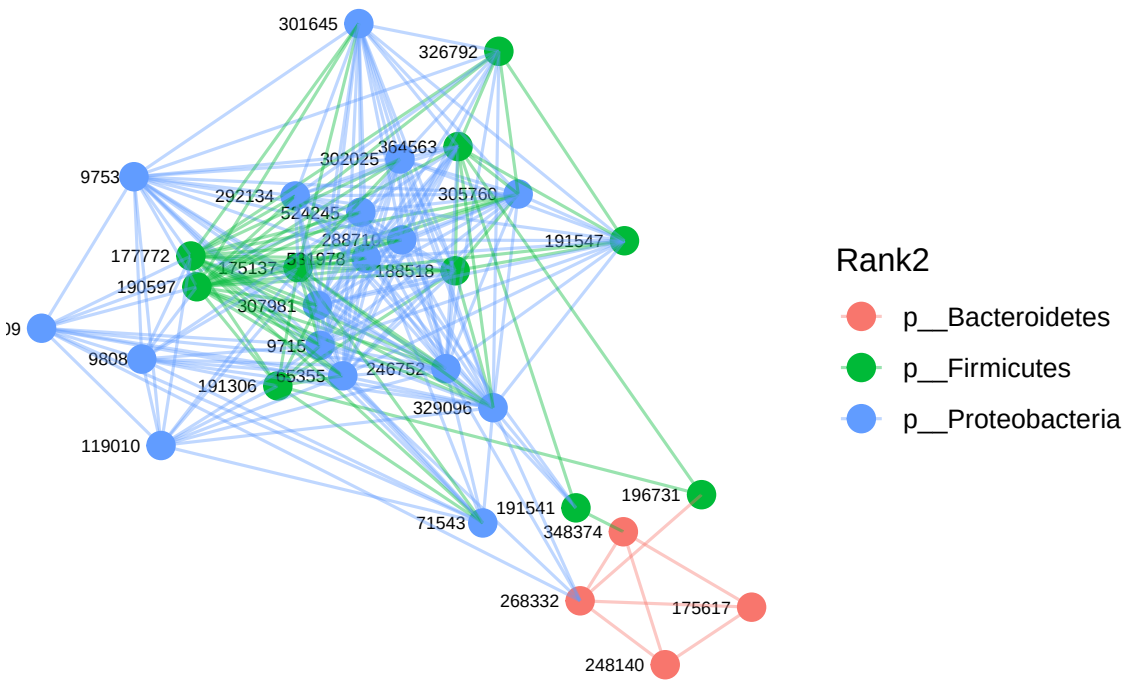

(c) Meta-variables graph with 31 clusters

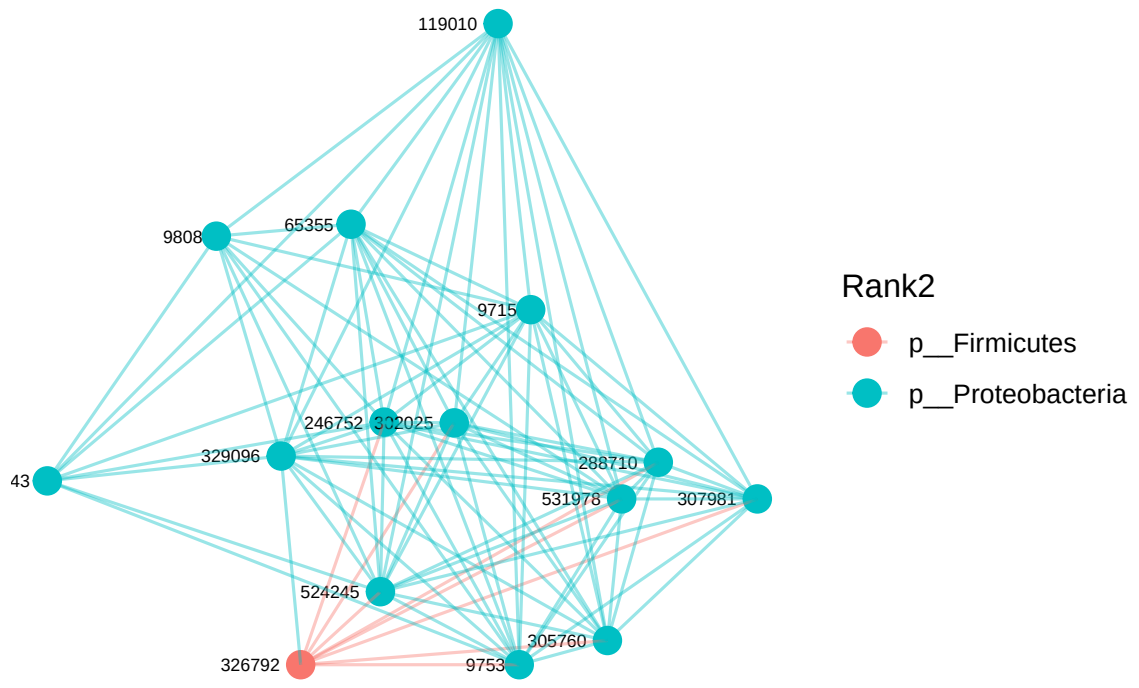

(d) Meta-variables graph with 15 clusters

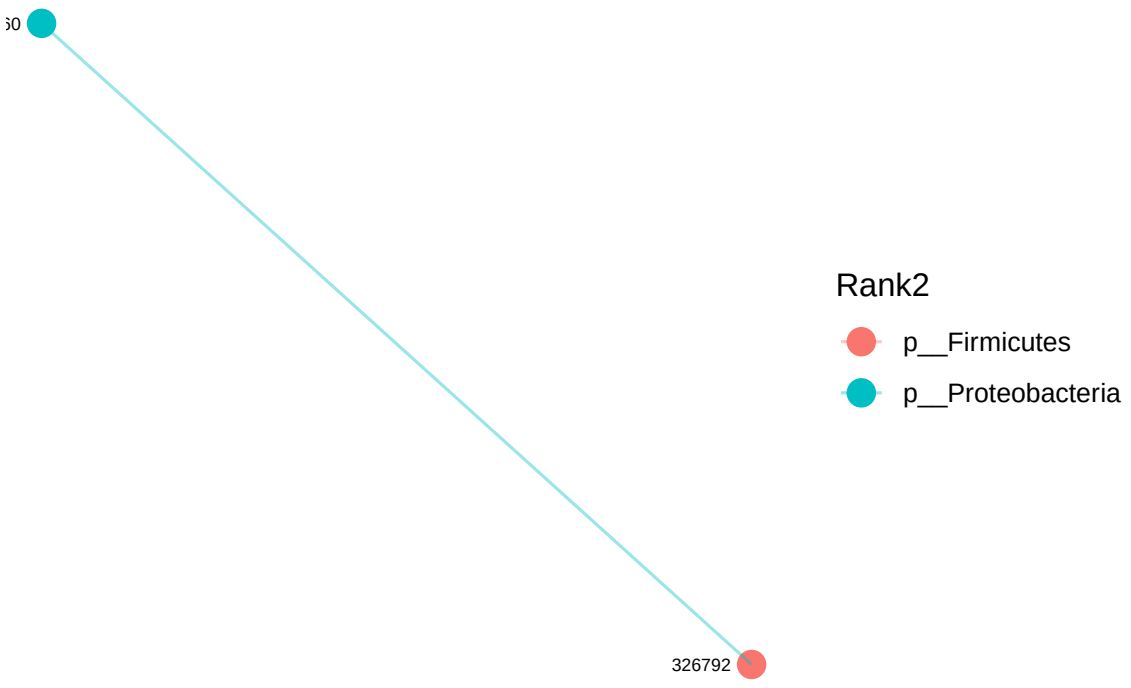

(e) Meta-variables graph with 2 clusters

Figure 11: Estimated graphs at multiple levels of granularity. The first graph shows a network inferred when $\lambda_2 = 0$.The number of clusters is equal to the number of OTUs. Increasing the fusion penalty makes it possible to uncover graphs built on the representative variable of each cluster. OTUs are colored according to their phylum taxonomic classifier. The number of clusters is computed from the regression vectors with a fixed fusion threshold.

phylum. The cluster 1 also contains Proteobacteria OTUs, so those identified in cluster 2 might share more intimate characteristics.

| | taxonomic.classification | | | | | |
|---|---|---|---|---|---|---|
| clusters | Actinobacteria | Bacteroidetes | Firmicutes | Proteobacteria | Tenericutes | Verrucomicrobia |
| 1 | 2 | 27 | 76 | 13 | 1 | 1 |
| 2 | 0 | 0 | 0 | 7 | 0 | 0 |

FALSE

Adjusted Rand indices are not calculated for comparisons as the unitary weights in the convex clustering problem can be suboptimal. The abundance of OTUs belonging to cluster 1, mainly composed of Bacteroidetes and Firmicutes phyla, is seemingly dependent on the abundance of OTUS in cluster 2, i.e., Proteobacteria phylum.

## 5.2 Application to methylation and transcriptomic genotypes in poplar

Next, we investigate interactions between European poplar genotypes for transcriptomic and DNA methylation data extracted from the Evolutionary and functional impact of EPIgenetic variation in forest TREEs project (EPITREE, Maury et al. 2019). The analysis was purposefully applied to the samples and not the genes in order to highlight the MGLasso clustering performance and show some potential relationships between DNA methylation and gene expression levels for some genotypes.

Poplar (*Populus*) is often used as a model tree for the study of drought response. Natural populations of black poplars (*Populus nigra*) have been planted in common gardens in France, Italy, and Germany (see Figure **??**) with control on some environmental variables such as water availability (Sow et al. 2018). The poplar has economic importance and is one of the most endangered species as a result of global climate change. The drought response can be studied via DNA methylation, which is a necessary process in plant development and response to environmental variations (Amaral et al. 2020). It consists of the addition of a Methyl group to a cytosine (C) in the genome and occurs in three contexts (CG, CHG, and CHH, where $\text{H} \in \{A, C, T\}$). Methylation can be measured on two regions of the gene. Methylation in promoters is linked to gene silencing, and methylation in the body of the gene can be related to tissue-specific expression or alternative splicing (Sow 2019).

The collected DNA methylation and expression data are counts data. Details on the plant material and experimental design can be found in Sow (2019) and Chateigner et al. (2020). The transcriptomic data were measured via RNA-Seq and normalized using Trimmed Mean of M-Values combined with a Best linear unbiased predictor (BLUP) correction as described in Chateigner et al. (2020). The methylation data were measured through whole-genome bisulfite sequencing (WGBS) and are normalized via the read per density approach then passed to a logarithm function $log_2(x + 1)$ with $x \in \mathbb{R}$. For each one of the 10 populations (see Figure **??**), DNA methylation in CG, CHG, and CHH contexts for promoters and gene-body and RNA sequencing data are observed on genotypes. A mean measure is computed from two replicates per population. The analysis has been restricted to a set of 151 target genes which explains the most variability in the omics data and the subsequent number of samples from different omic variables, which is 70.

The MGLasso model is fitted with fusion penalty values chosen in $[0, 30.94]$ and a Lasso penalty $\lambda_1$ parameter chosen via the StARS approach with threshold 0.05. In the resulting clustering path (see Figure 13), we can identify three distinct and coherent clusters, which are samples corresponding to gene expression genotypes, gene-body methylation samples, and gene promoter samples.

The results of the MGLasso can also be represented in the expanded way where meta-variables are not computed from clusters. In Figure 14, a focus is put on the effect of the fusion penalty. Clusters partitions are not presented. The higher the fusion penalty, variables are encouraged to share the

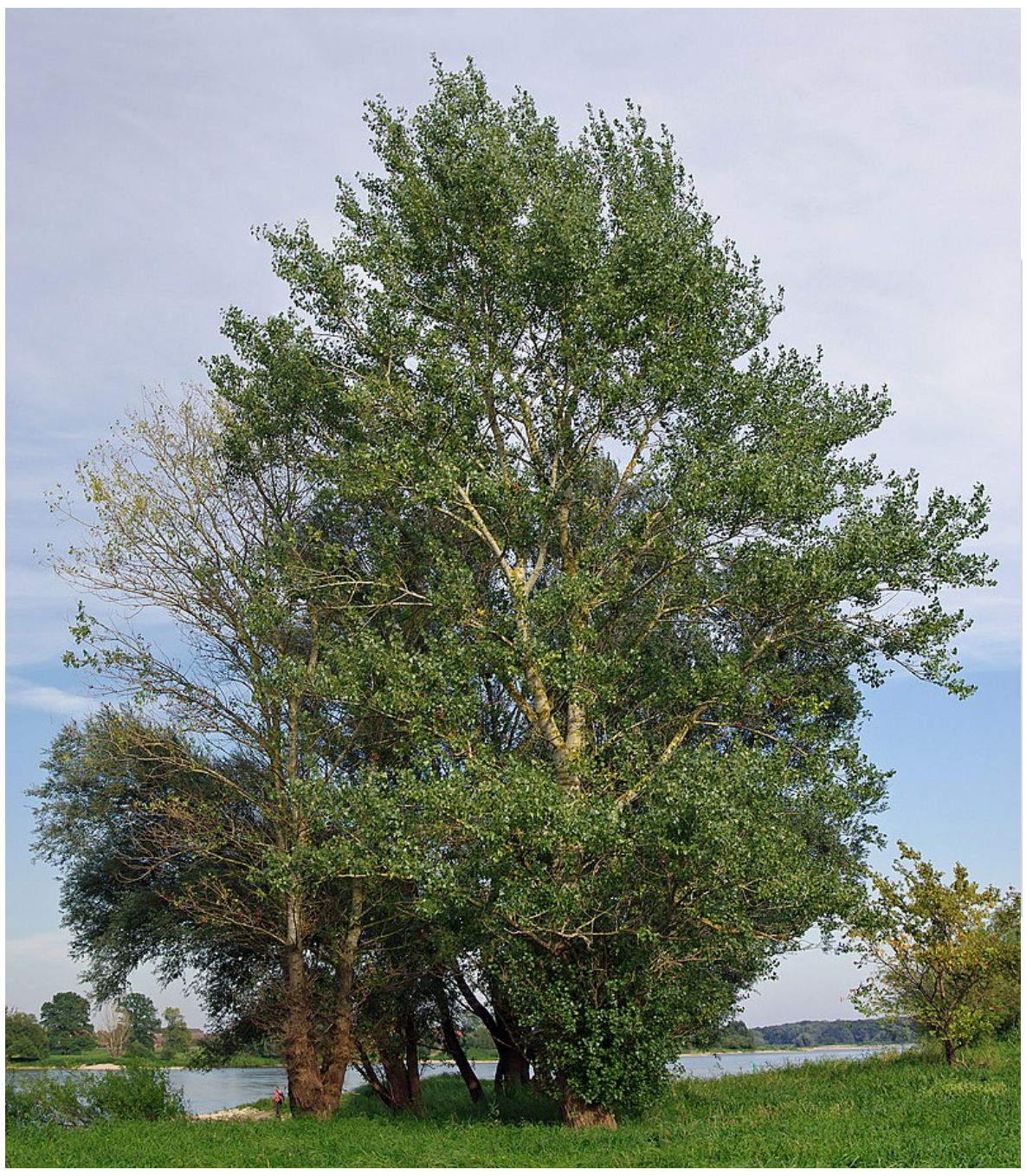

(a) Black poplar (C. Fischer Wikimedia)

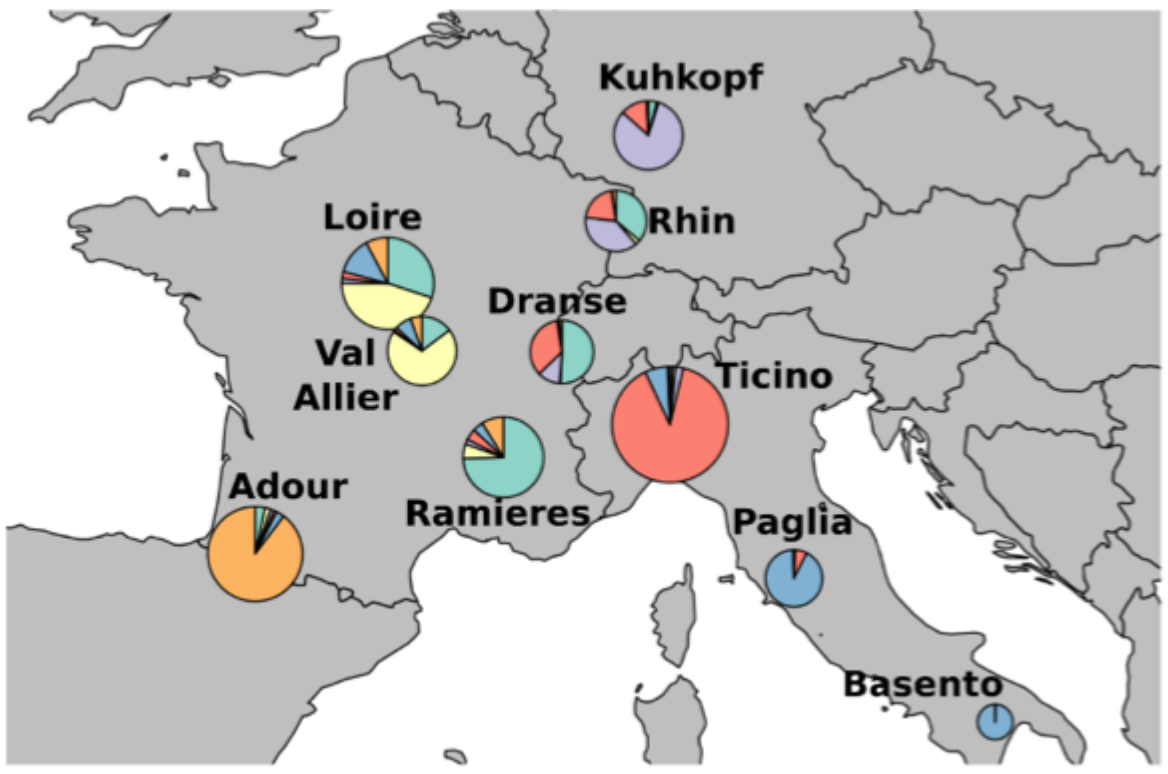

(b) Map of genotypes

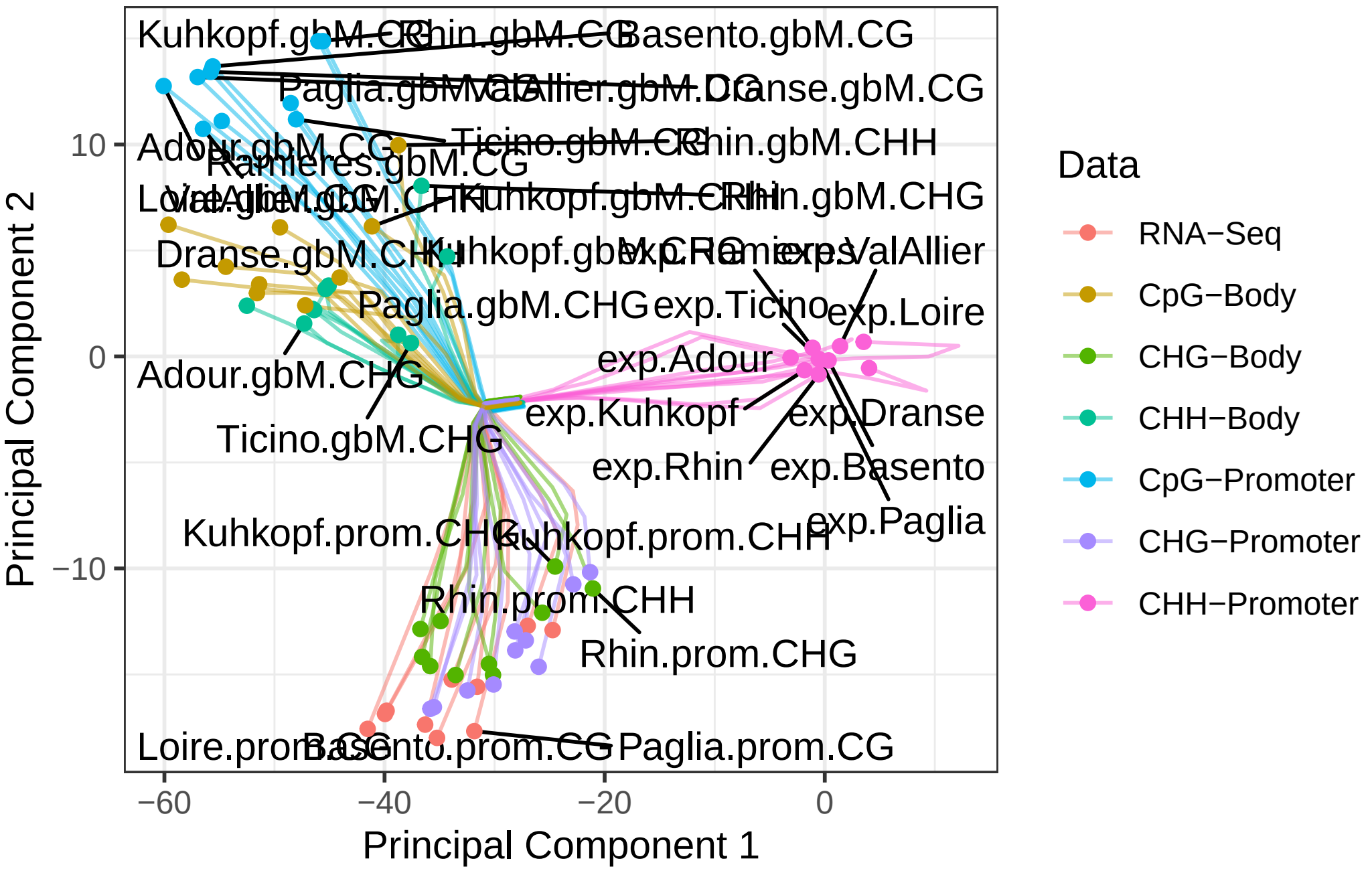

Figure 13: Clustering path of solutions on DNA methylation and transcriptomic samples. The figure shows 3 distinct clusters which correspond to omics data of different natures: transcriptomic (right), methylation on the promoter (bottom), and methylation on gene-body (top left).

same neighborhood structure. Note that an equivalent graph over meta-variables can be computed after choosing a fusion threshold as in Figure 11.

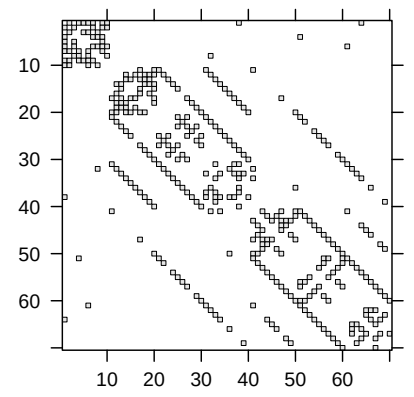

(a) Full graph with $\lambda_2 = 0$

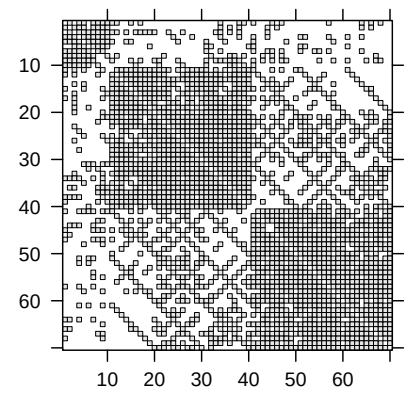

(b) Full graph with $\lambda_2 = 1.63$

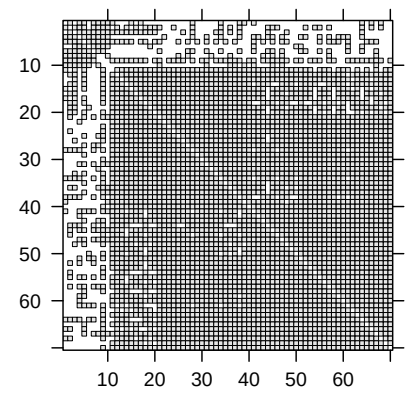

(c) Full graph with $\lambda_2 = 3.26$

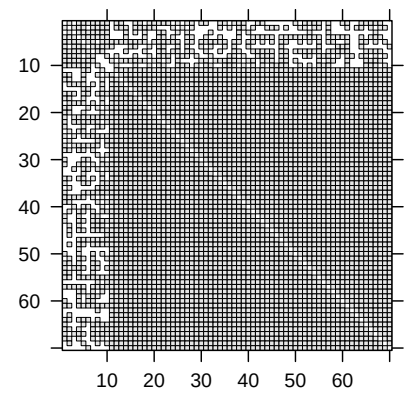

(d) Full graph with $\lambda_2 = 4.89$

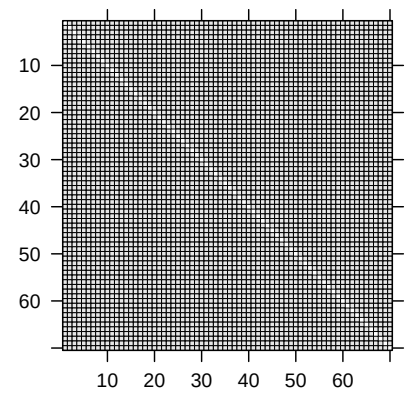

(e) Full graph with $\lambda_2 = 30.94$

Figure 14: Adjacency matrices for different fusion penalty parameters. The first graph shows the inferred network when no fusion penalty is added to the model. In that graph, the first block of size $10 \times 10$ variables corresponds to RNA-Seq samples. The second sparser block of size $30 \times 30$ corresponds to gene-body DNA methylation data in the three methylation contexts. The last sparse block of the same size corresponds to promoter methylation. The edge bands suggest a relationship between DNA methylation measurements that belong to the same context. For example, the Loire methylation sample in the CpG context is likely related to the Loire samples in the CHG and CHH contexts. The graphs also suggest some relationships between expression and methylation for some natural populations. As the merging penalty increases, the blocks corresponding to the three methylation contexts merge first, then follow the upper left block corresponding to the expression data. For $\lambda_2 = 30.94$, all natural populations merge into a single cluster and complete graph.

## 6 Conclusion

We proposed a new technique that combines Gaussian Graphical Model inference and hierarchical clustering called MGLasso. The method proceeds via convex optimization and minimizes the neighborhood selection objective penalized by a hybrid regularization combining a sparsity-inducing norm and a convex clustering penalty. We developed a complete numerical scheme to apply MGLasso in practice, with an optimization algorithm based on CONESTA and a model selection procedure. Our simulations results over synthetic and real datasets showed that MGLasso can perform better than GLasso in network support recovery in the presence of groups of correlated variables, and we illustrated the method with the analysis of microbial associations data and methylation mixed with transcriptomic data. The present work paves the way for future improvements: first, by incorporating prior knowledge through more flexible weighted regularization; second, by studying the theoretical properties of the method in terms of statistical guarantees for the MGLasso estimator. Moreover, the node-wise regression approach on which our method is based can be extended to a broader family of non-Gaussian distributions belonging to the exponential family as outlined by E. Yang et al. (2012). Our MGLasso approach can be easily extended to non-Gaussian distributions belonging to the exponential family and mixed graphical models.

## Appendix

The scripts to reproduce the simulations are available at https://github.com/computorg/published-202306-sanou-multiscale_glasso/tree/main/scripts/simulation-experiments.

## Acknowledgments

The authors would like to thank the Editors and referees for comments that led to substantial improvements in the manuscript.

## Session information

```r
sessionInfo()
```

```
R version 4.5.0 (2025-04-11)
Platform: x86_64-pc-linux-gnu
Running under: Ubuntu 24.04.3 LTS

Matrix products: default
BLAS:   /usr/lib/x86_64-linux-gnu/blas/libblas.so.3.12.0
LAPACK: /usr/lib/x86_64-linux-gnu/lapack/liblapack.so.3.12.0  LAPACK version 3.12.0

locale:
 [1] LC_CTYPE=C.UTF-8       LC_NUMERIC=C           LC_TIME=C.UTF-8
 [4] LC_COLLATE=C.UTF-8     LC_MONETARY=C.UTF-8    LC_MESSAGES=C.UTF-8
 [7] LC_PAPER=C.UTF-8       LC_NAME=C              LC_ADDRESS=C
[10] LC_TELEPHONE=C         LC_MEASUREMENT=C.UTF-8 LC_IDENTIFICATION=C

time zone: Etc/UTC
tzcode source: system (glibc)

attached base packages:
[1] stats     graphics  grDevices datasets  utils     methods   base

other attached packages:
[1] phyloseq_1.52.0  igraph_2.1.4     ggrepel_0.9.6    colorspace_2.1-1
[5] SpiecEasi_1.1.3  ghibli_0.3.4     ggplot2_3.5.2    mglasso_0.1.3

loaded via a namespace (and not attached):
  [1] simone_1.0-4          gld_2.6.7             permute_0.9-7
  [4] readxl_1.4.5          rlang_1.1.6           magrittr_2.0.3
  [7] ade4_1.7-23           e1071_1.7-16          compiler_4.5.0
 [10] mgcv_1.9-3            png_0.1-8             vctrs_0.6.5
 [13] reshape2_1.4.4        stringr_1.5.1         pkgconfig_2.0.3
 [16] shape_1.4.6.1         crayon_1.5.3          fastmap_1.2.0
 [19] XVector_0.48.0        labeling_0.4.3        rmarkdown_2.29
 [22] tzdb_0.5.0            haven_2.5.5           UCSC.utils_1.4.0
 [25] xfun_0.52             glmnet_4.1-9          GenomeInfoDb_1.44.0
 [28] jsonlite_2.0.0        biomformat_1.36.0     rhdf5filters_1.20.0
 [31] Rhdf5lib_1.30.0       huge_1.3.5            VGAM_1.1-13
```

```
 [34] cluster_2.1.8.1         parallel_4.5.0          DescTools_0.99.60
 [37] R6_2.6.1                tables_0.9.31           stringi_1.8.7
 [40] RColorBrewer_1.1-3      reticulate_1.42.0       boot_1.3-31
 [43] cellranger_1.1.0        Rcpp_1.0.14             iterators_1.0.14
 [46] knitr_1.50              readr_2.1.5             IRanges_2.42.0
 [49] Matrix_1.7-3            splines_4.5.0           tidyselect_1.2.1
 [52] rstudioapi_0.17.1       yaml_2.3.10             vegan_2.6-10
 [55] codetools_0.2-20        lattice_0.22-7          tibble_3.2.1
 [58] plyr_1.8.9              Biobase_2.68.0          withr_3.0.2
 [61] evaluate_1.0.3          survival_3.8-3          proxy_0.4-27
 [64] capushe_1.1.2           Biostrings_2.76.0       pillar_1.10.2
 [67] BiocManager_1.30.25     renv_1.1.4              foreach_1.5.2
 [70] stats4_4.5.0            generics_0.1.4          hms_1.1.3
 [73] S4Vectors_0.46.0        scales_1.4.0            rootSolve_1.8.2.4
 [76] class_7.3-23            glue_1.8.0              pulsar_0.3.11
 [79] lmom_3.2                tools_4.5.0             data.table_1.17.4
 [82] forcats_1.0.0           Exact_3.3               fs_1.6.6
 [85] mvtnorm_1.3-3           rhdf5_2.52.0            grid_4.5.0
 [88] ape_5.8-1               nlme_3.1-168            GenomeInfoDbData_1.2.14
 [91] cli_3.6.5               expm_1.0-0              dplyr_1.1.4
 [94] gtable_0.3.6            digest_0.6.37           prismatic_1.1.2
 [97] BiocGenerics_0.54.0     farver_2.1.2            multtest_2.64.0
[100] htmltools_0.5.8.1       blockmodels_1.1.5       lifecycle_1.0.4
[103] httr_1.4.7              MASS_7.3-65
```